\documentclass{article} 
\usepackage{iclr2026_conference,times}

\usepackage{amsmath,amsfonts,bm}

\def\eqref#1{equation~\ref{#1}}

\def\1{\bm{1}}

\DeclareMathAlphabet{\mathsfit}{\encodingdefault}{\sfdefault}{m}{sl}
\SetMathAlphabet{\mathsfit}{bold}{\encodingdefault}{\sfdefault}{bx}{n}

\usepackage{hyperref}
\hypersetup{
    colorlinks,
    linkcolor={red!50!black},
    citecolor={blue!50!black},
    urlcolor={blue!80!black}
}
\usepackage{cleveref}
\AddToHook{cmd/appendix/before}{%
    \crefalias{section}{appendix}%
    \crefalias{subsection}{appendix}
}
\usepackage{url}
\usepackage[textsize=tiny]{todonotes}

\usepackage{booktabs}
\usepackage[table]{xcolor}
\usepackage{vcell}
\usepackage{subcaption}
\usepackage{graphicx}
\usepackage{amsmath}
\usepackage{float}
\usepackage{amssymb}
\usepackage{tikz}
\usepackage{xspace}
\usepackage{bm}
\usepackage{wrapfig}
\usepackage{multirow}
\usepackage{multicol}
\usepackage{duckuments}
\usepackage{graphicx}
\usepackage{pifont} %
\usepackage{listings}
\usepackage{enumitem}

\newtoggle{final}
\toggletrue{final}
\iftoggle{final}{}{\setlength{\marginparwidth}{2cm}}
\newcommand{\pw}[1]{\iftoggle{final}{#1}{{\color{blue} #1}}}
\newcommand{\hjs}[1]{\iftoggle{final}{#1}{{\color{red} #1}}}

\newcommand{\revise}[1]{{\color{black} #1}}

\newcommand{\method}{CLAP\xspace}

\newcommand{\yes}{\color{blue}{\ding{51}}}
\newcommand{\no}{\color{red}{\ding{55}}}

\newcommand{\rpmh}{\huge \raisebox{.2ex}{$\scriptstyle\pm~$}}

\newcommand{\tb}[1]{\textbf{#1}}

\title{Generalizable Coarse-to-Fine Robot Manipulation via Language-Aligned 3D Keypoints}

\makeatletter
\def\@fnsymbol#1{\ensuremath{\ifcase#1\or \dag\or \ddag\or \S\or \P\or \|\or **\or \dagger\dagger\or \ddagger\ddagger \else\@ctrerr\fi}}
\makeatother

\author{Jianshu Hu, Lidi Wang, Shujia Li, Yunpeng Jiang, \& Yutong Ban\thanks{indicates corresponding author}
\\
Global College\\
Shanghai Jiao Tong University\\
\texttt{\{hjs1998, lidiwang, sjtu7264970, jyp9961, yban\}@sjtu.edu.cn} \\
\And
Xiao Li \\
School of Mechanical Engineering \\
Shanghai Jiao Tong University \\
\texttt{\{sjtu\_lixiao\}@sjtu.edu.cn} \\
\And
Paul Weng$^{\dag}$ \\
Digital Innovation Research Center \\
Duke Kunshan University \\
\texttt{\{paul.weng\}@dukekunshan.edu.cn} \\
}

\iclrfinalcopy 
\begin{document}

\maketitle

\begin{abstract}
Hierarchical coarse-to-fine policy, where a coarse branch predicts a region of interest to guide a fine-grained action predictor, has demonstrated significant potential in robotic 3D manipulation tasks by especially enhancing sample efficiency and enabling more precise manipulation.
However, even augmented with pre-trained models, these hierarchical policies still suffer from generalization issues.
To enhance generalization to novel instructions and environment variations, we propose Coarse-to-fine Language-Aligned manipulation Policy (CLAP), a framework that integrates three key components: 1) task decomposition, 2) VLM fine-tuning for 3D keypoint prediction, and 3) 3D-aware representation.
Through comprehensive experiments in simulation and on a real robot, we demonstrate its superior generalization capability.
Specifically, on GemBench, a benchmark designed for evaluating generalization, our approach achieves a 12\% higher average success rate than the SOTA method while using only 1/5 of the training trajectories.
In real-world experiments, our policy, trained on only 10 demonstrations, successfully generalizes to novel instructions and environments.
\end{abstract}

\section{Introduction}


Robot learning, especially via imitation learning, has demonstrated promising success in enabling robots to solve complex 3D manipulation tasks \citep{pi_05, rdt-1b}.
However, scaling these methods to a broader range of real-world applications (e.g., industrial, service, or home robotics) requires enhancing both (G1) their generalization to environment variations, and (G2) their skill compositional generalization.
Indeed, G1 is necessary, because deployed robots need to be able to operate in new settings (e.g., object or background variation), while G2 is highly desirable, so that trained robots can tackled new tasks by composing previously-learned skills.
To achieve G1 and G2, the robot needs to be endowed with a combination of capabilities, such as scene understanding, reasoning or planning, and high-precision manipulation, exploiting preferably sample efficient techniques, since robotics data is costly to collect.

In this paper, we focus on one type of 3D manipulation policies, called \emph{coarse-to-fine} policies \citep{HSA, c2f_bc, coarse-to-fine-affordance, rvt2, act3D, FP2AT}, because they achieve superior precision in manipulation tasks while enjoying strong sample efficiency.
These policies process 3D observations (or 3D scene representations) using a hierarchical architecture whose higher-level coarse branch identifies a region of interest for the lower-level fine-grained branch to focus on and predict a final action.
Typically, the coarse branch is trained to predict a 3D keypoint, which serves as the center for cropping and zooming into the original 3D observations.
To help with visual understanding and to some extent spatial reasoning, recent work \citep{bridgevla, sam2act} has extended this approach to exploit pre-trained \pw{models---}Vision-Language Models (VLMs) \citep{paligemma} \hjs{or \pw{visual foundation models} \citep{sam2}}.
However, the performance of these \pw{obtained} methods is still limited in terms of generalization capability (G1 and especially G2), \pw{indicating} that their \hjs{scene understanding and reasoning} capabilities are actually still rudimentary. 
Our experimental study suggests that this is primarily due to a \pw{combination of various issues (depending on the method), such as} 
domain shift between pre-training and robotic \pw{images}, 
inadequacy of pre-trained models to predict 3D \pw{keypoint}, 
poor adaptation to object variations, \pw{or} 
under-exploitation of the planning ability of VLMs.

To address these limitations and issues, we propose Coarse-to-fine Language-Aligned manipulation Policy (\textbf{\method}), a novel coarse-to-fine 3D manipulation policy.
In contrast to previous coarse-to-fine policies, \method includes a novel architecture for the higher-level branch, which we name \emph{coarse task planner}, and a novel implementation of the lower-level fine-grained action predictor, both leveraging pre-trained models.

The coarse task planner, implemented as a VLM, is introduced to play the additional role of task planning. Before the usual 3D keypoint prediction, it decomposes a task into step-wise language instructions, representing basic skills. 
This change allows both 3D keypoint and action predictions to depend on step-wise instructions instead of the whole task description, which promotes skill compositional generalization (G2).
The training of this coarse task planner consists of three parts to reinforce its \hjs{scene understanding and reasoning} capabilities. 
First, the pre-trained VLM fine-tuned on language plans of different tasks to directly improve compositional reasoning.
Second, it is specialized for 3D keypoint prediction by fine-tuning it to perform a sequential reasoning process: first localizing task-related objects, then generating the step instruction, and finally predicting a corresponding 3D keypoint.
Finally, to further boost \pw{its} scene understanding capability, the VLM is further fine-tuned with an auxiliary task of 3D object detection, using an additional dataset of object positions.
Together, these components form a comprehensive pipeline that significantly enhances the generalization ability of the coarse-to-fine policy to object variations (G1) and novel tasks (G2).

The fine-grained action predictor takes as input both the step instruction and the multi-view RGB-D images and outputs an action.
It is implemented with specialized pre-trained models to improve sample efficiency and increase its precision during manipulation.
More specifically, step instruction and RGB images are processed using a pre-trained visual-language encoder, ensuring the two modalities are well-aligned.
The depth information is processed by a dedicated encoder and augmented with 3D position embeddings to help better align 3D and 2D image information.
All the obtained embeddings\pw{, which we call 3D-aware representation,} are fused via a 
\hjs{Multi-View Transformer \citep{rvt}}
to predict the final actions.

To evaluate the performance of our method, we run experiments in both simulation and real-world. 
For simulation, we use GemBench \citep{gembench}, a benchmark specifically designed to assess the generalization ability of multi-task language-conditioned policies across varying difficulty levels.
Our approach outperforms the state-of-the-art method, achieving a 12\% higher average success rate with only 1/5 of the training trajectories.
In real-world experiments, our method demonstrate strong generalization ability to novel tasks and object variations with only 10 demonstrations per task. 

\begin{figure}
    \centering
    \includegraphics[width=0.7\linewidth]{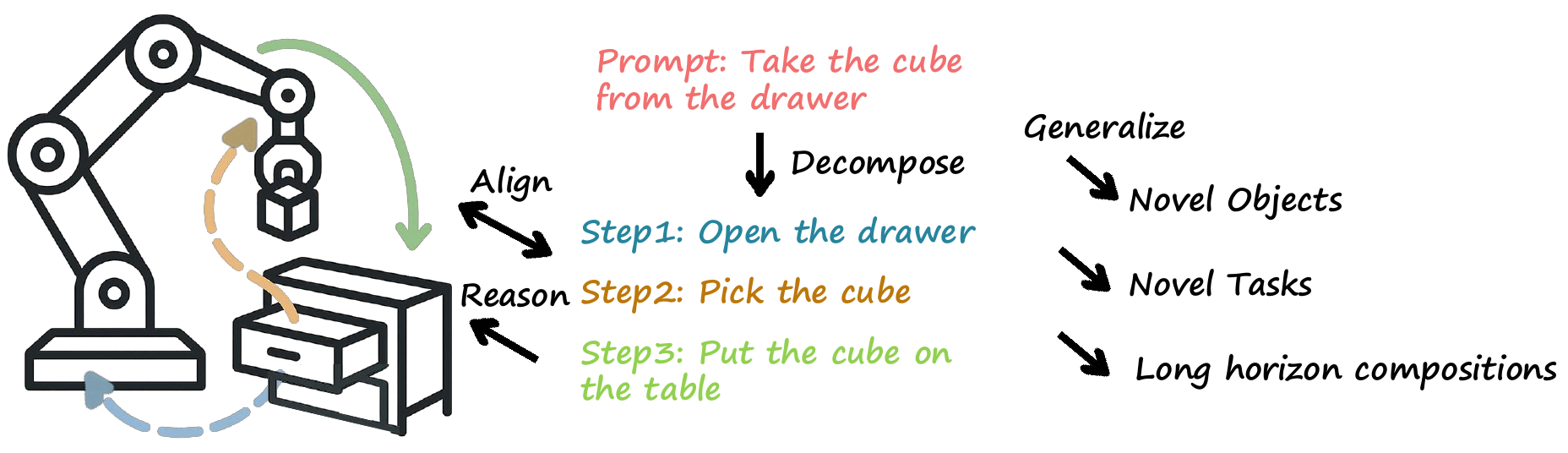}
    \caption{\tb{Intuition of \method.} Our method achieves strong generalization ability by decomposing tasks into step-wise language instructions, each aligned with a 3D keypoint.}
    \label{fig:teaser}
\end{figure}

\paragraph{Contributions}
\begin{enumerate}
    \item 
    We introduce a novel coarse-to-fine 3D manipulation policy, as shown in \Cref{fig:teaser}, \pw{with two main innovations: (1)} tasks are decomposed into step-wise language instructions to promote compositional generalization ability\pw{; (2) action inference is performed via a reasoning step to improve generalization to object variations.}
    \item 
    We design a finetuning pipeline that effectively adapts a pre-trained VLM to 3D keypoint prediction and incorporate a \pw{3D-aware representation} in the fine-grained action predictor\pw{, overcoming the issues observed in previous methods.}
    \item 
    \pw{Empirical} evaluations in simulation and on a real robot demonstrate state-of-the-art performance in both robustness to \pw{visual and object changes} and generalization to unseen tasks.
\end{enumerate}

\section{Related Work}
In this section, we discuss the related works in the field, including vision-language-action models, 3D manipulation policies, and coarse-to-fine policies.

\paragraph{Vision-Language-Action (VLA) models}
\pw{Training VLMs \citep{chatgpt4o, paligemma, Qwen2.5-VL, deepseekv3} on vast internet-scale image-text corpora has led to} remarkable capabilities in image understanding, excelling at tasks like image classification, object detection\pw{,} and visual question answering tasks.
However, applying \pw{a similar training strategy} directly to robotics presents a challenge \pw{due to} the \pw{relatively scarce} robot trajectory data.
A prominent solution is to transfer the knowledge from pre-trained VLMs by fine-tuning them on robot data.
This approach is the foundation for recent VLA models \citep{palm-e, rt2, openvla, octo, pi_05, tinyvla, smolvla, rdt-1b, dexvla, cogact, gr3, geminirobotics, gr00t}, which are fine-tuned on \pw{large} diverse datasets of robot trajectories.
Such extensive training \pw{strengthens} generalization to novel objects, environments, and \pw{tasks}.
However, \pw{since} they commonly use multi-view \pw{2D} images as visual input\pw{,} learning to reason \pw{in} 3D space from 2D images alone is data-intensive.
This leads to sample inefficiency and low success rates on some tasks.
Recent work ha\pw{s} sought to more explicitly incorporate 3D information \citep{pointvla, spatialvla, 3dvla} or introduce Chain of Thought \citep{embodiedgpt, robotic_cot, CoTVLA} \pw{to enhance the 3D} reasoning ability.
However these directions remain relatively underexplored within the VLA paradigm.
Our method, which fine-tunes a pretrained VLM as a coarse task planner \hjs{and predicts the final action with a fine-grained action predictor}, can also be viewed as a VLA model.
\pw{In contrast to other VLA approaches, we propose specific training and inference techniques to better align pre-trained VLMs to 3D manipulation, further enhancing generalization (G1-G2) while retaining the sample efficiency inherent to hierarchical coarse-to-fine policies.}

\paragraph{3D Manipulation Policy}
3D manipulation policies \citep{peract, act3D, lift3D, spa, gendp, dp3, rise, rvt2,  3ddiffuseractor, GravMAD, DeCo, sam2act, bridgevla, gembench} \pw{directly work with 3D inputs and outputs.
They generally include structured architectures that construct a 3D representation of the scene,} 
\pw{leading to higher sample efficiency and better generalization to new camera viewpoints.}
For example, PerAct \citep{peract} explicitly represents the scene with a voxel representation.
\citet{act3D} and \citet{3ddiffuseractor} process \pw{RGB} image\pw{s} with pre-trained image encoder and lift 2D features to 3D by aggregating \pw{with} depth information.
An alternative approach \citep{lift3D, rvt2} is to project point clouds into canonical virtual view\pw{s} and use the resulting multi-view image\pw{s} as input.
\pw{Explicitly exploiting 3D information allows these models to achieve high success rates with much less training data, which can be further reduced by enforcing a hierarchical structure like in coarse-to-fine policies.}

\paragraph{Coarse-to-fine Policies}
\citet{first_attention_focus, HSA} first propose this coarse-to-fine scheme for pick-and-place tasks.
Subsequent work has considered more general tasks and explored various 3D representations, such as voxel observations \citep{c2f_bc,FP2AT}, 3D feature fields \citep{act3D}, and multi-view images \citep{rvt2}.
\citet{coarse-to-fine-affordance} \pw{apply} the coarse-to-fine architecture to handle noisy point cloud\pw{s}.
Among these, Robotic View Transformer 2 (RVT2) \citep{rvt2} \pw{is an effective language-conditioned multi-task policy, demonstrating} strong performance in both training and inference efficiency by using multi-view images projected from canonical views.
However, RVT2 is trained from scratch\pw{,} which limits its generalization ability to visual perturbations and task variations.
Subsequent efforts built upon this work have sought to \pw{overcome} these limitations.
Existing works \citep{bridgevla, sam2act} have attempted to enhance generalization through strategies such as: pre-training on object detection datasets \citep{robopoint} or integrating encoders from powerful visual foundation models like Segment Anything Model 2 \citep{sam2}.
In contrast, we achieve this by introducing a \pw{novel architecture,} where tasks \pw{are decomposed} into step-wise language instructions \pw{for skill compositional generalization} and design \pw{a specific training and inference} pipeline to leverage pre-trained models in both coarse and fine-grained branches.

\section{Background}
In this section, we first briefly \pw{recall} the \pw{multi-task imitation learning set-up, introduce coarse-to-fine policy,} and then present Robotic-View-Transformer 2 (RVT2) \citep{rvt2}, a state-of-the-art coarse-to-fine policy that serves as the foundation for our method.

\pw{In multi-task imitation learning,} a dataset $\mathcal{D}=\{(\tau_i, L_i)\mid i={1,...,N}\}$ \pw{is available for pairs} of robot demonstrations ${\tau_i}$ and task description $L_i$.
\pw{A robot} demonstration is a trajectory $\tau_i = (o_0,a_0,o_1,a_1, ...)$ containing a sequence of observations $o_t$ and corresponding expert actions $a_t$.
\pw{Observations include}
multi-view RGB-D images and \hjs{gripper status} \pw{indicating whether it is close or open}.
\pw{Actions denote} the state of the end-effector\pw{,} which contains the 3D position $p_t= (x_{t}, y_{t}, z_{t})$ of the gripper, the orientation of the gripper and a gripper status.

\pw{A c}oarse-to-fine policy contains a coarse branch and a fine-grained branch, where the coarse branch predicts a 3D keypoint as the center to zoom in the 3D observation and the fine-grained branch uses the refined observation to predict the target action.
\pw{Such policy is trained according to the} key-frame based imitation learning framework \citep{key-frame, peract, rvt2}.
Specifically, key-frames \pw{identifies timesteps} \pw{in a trajectory when an important action,} like grasping or placing\pw{, occurs}.
\pw{In practice, they are usually} heuristically defined for each trajectory.
With these key-frames, \pw{a} trajectory is segmented into $K$ \pw{sub}sequences of observations and actions $(o_0,a_0,...,o_{t_1},a_{t_1}), ...,(o_{t_{K-1}+1},a_{t_{K-1}+1},...,o_{t_K},a_{t_K})$, where the $k^{th}$ key-frame occurs at time step $t_k$\pw{, from which} we can extract a sequence of key-frame actions $(a_{t_1}, ...,a_{t_K})$.
\pw{In this framework, t}he goal is to train a policy $\pi$ to predict the key-frame action \pw{$a_{t_k}$} at the closest next key-frame \pw{of timestep $t_k$} given \pw{an} observation \pw{$o_t$} and a task description \pw{$L_i$}:
\begin{equation}
    \pi(o_t, L_i) \rightarrow a_{t_k} \mbox{ for }
    t_{k-1}\leq t < t_k\,.
\end{equation}
\hjs{The predicted actions are executed by a motion planner\pw{,} which moves the robot to the desired state}\pw{, generating thus the intermediate actions in a trajectories.}
\hjs{In coarse-to-fine policies,
the 3D position $p_{t_k}$ \pw{output by the coarse branch for} the next key-frame is typically used as the 3D keypoint to zoom \pw{into} the observation \pw{for the action predictor}.}

In RVT2, multi-view RGB-D images 
are first aggregated into a point cloud, which is then projected into three canonical views: front, left and top.
These three views are orthogonal to each other, which allows a mapping between pixel positions in these views and a 3D position in the scene.
\hjs{Each pixel in the projected images contains 3-channel RGB values, 1-channel depth value and its corresponding 3D position in the global coordinate.}
In the coarse branch, the projected images are tokenized using convolutional layers while the task description and robot states (e.g., gripper status) are encoded by a pre-trained language encoder and a
\pw{trainable} Multi-Layer-Perceptron respectively. 
All these tokenized features are fused via 
\hjs{\pw{Multi}-View Transformer \citep{rvt}}. 
The image tokens in the output of the transformer are then processed by upsampling layers to predict heatmaps, from which a 3D keypoint is extracted.
The keypoint from the coarse branch is used to zoom in and crop the point cloud while the cropped region is again projected into the canonical views.
The refined observations along with the same task description are processed by the fine-grained branch\pw{, implemented as} \hjs{\pw{another} multi-view transformer with different weights}\pw{,}
to predict the final actions.
While RVT2 achieves strong sample efficiency
and enables precise manipulation via \hjs{projections} \pw{to canonical views and} its coarse-to-fine architecture, it is trained from scratch and \pw{therefore does not leverage recent pretrained large models.
In addition, its architecture design} does not fully exploit common skills among tasks.
\pw{As a result, it suffers from deficient} generalization to \pw{visual changes, object variations, and} novel tasks.

\section{Method}
\label{Sec:method}
\begin{figure}
    \centering
    \includegraphics[width=\linewidth]{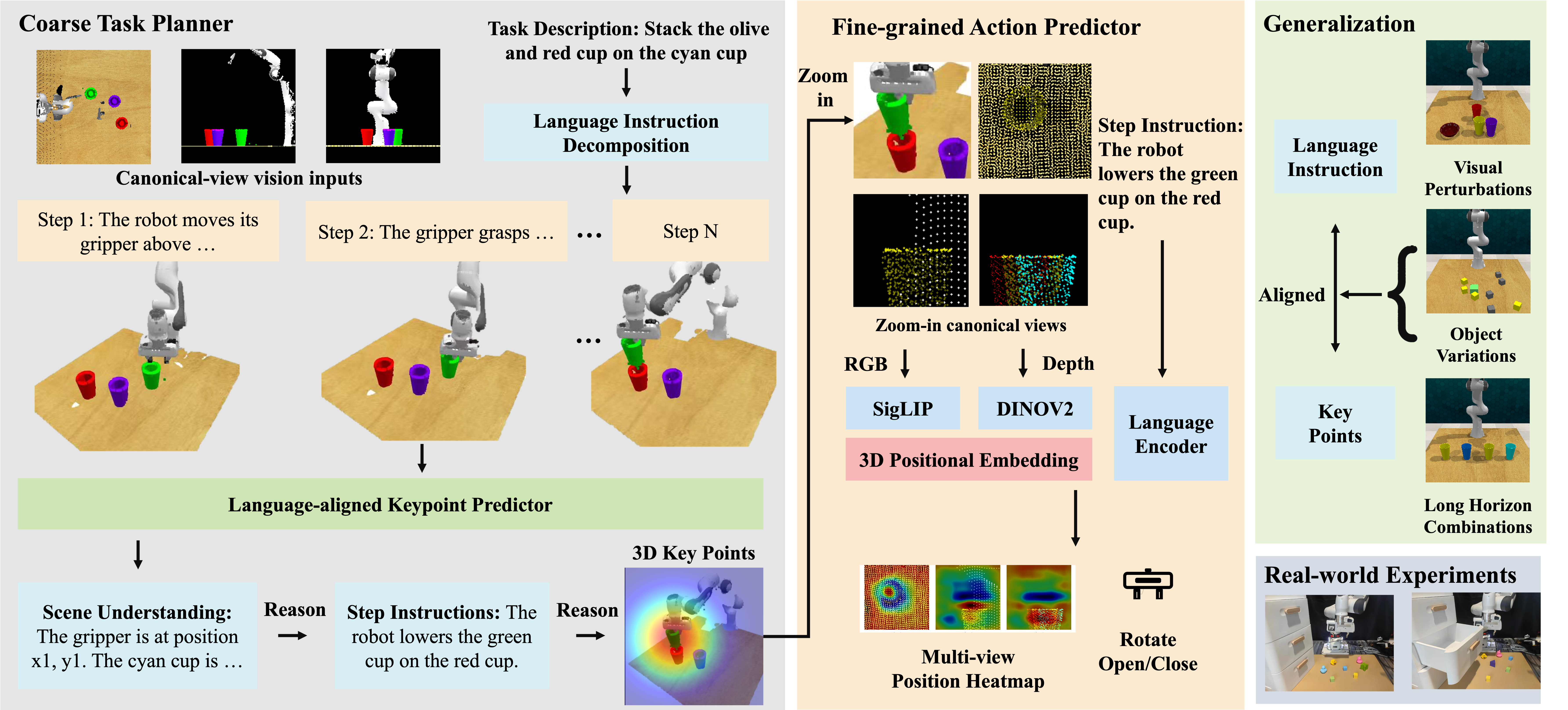}
    \caption{\tb{Overview of \method.}
    We propose a novel coarse-to-fine 3D manipulation policy, comprising of a coarse task planner and a fine-grained action predictor.
    The coarse task planner reasons about the task plans and the positions of task-related objects to generate language-aligned 3D keypoints.
    The fine-grained action predictor fuses the corresponding step instruction with a 3D-aware visual representation from refined observations to predict the final action.}
    \label{fig:method}
\end{figure}

Our hierarchical policy consists of a coarse task planner and a fine-grained action predictor, as shown in \Cref{fig:method}.
To promote compositional generalization, we first decompose tasks into step-wise language instructions, each describing the motion of the robot between two consecutive key-frames.
This enables the coarse task planner to perform language-guided planning while allowing the fine-grained action predictor to learn and reuse common skills across tasks.
Moreover, we adapt a pre-trained VLM to 3D keypoint prediction in the coarse task planner, by finetuning it with a sequential reasoning procedure. 
\revise{The VLM predicts the pixel position in each view and we follow the prior work \citep{rvt2} to map them to a 3D position.}
The pretrained VLM is trained to first reason about the positions of task-related objects, then generate a step instruction and finally predict the corresponding 3D keypoints.
To enable zero-shot generalization ability to novel objects, we add an auxiliary task of 3D object detection by augmenting the training data with additional object positions dataset.
Finally, \pw{in contrast to RVT2,} \pw{the coarse and fined-grained branches are implemented with a different architecture, since they play different roles.
Thus, for the latter, we utilize} \pw{specialized} pre-trained \pw{visual foundation models} to construct a 3D-aware representation.
The design choices of both the coarse task planner and the fine-grained action predictor are detailed in the following sections.

\subsection{Coarse Task Planner}



Prior coarse-to-fine policies condition all actions within a trajectory on a single high-level task description, limiting compositional generalization.
To address this, we \pw{leverage a pre-trained VLM, denoted $f_\theta$, to decompose a high-level task description $L$} into step-wise language instructions $\mathcal{L}=(\ell_1,...,\ell_k,...,\ell_K)$, 
which naturally align with the key-frame based trajectory segmentation.
Each trajectory segment $(o_{t_{k-1}+1},a_{t_{k-1}+1},...,o_{t_k},a_{t_k})$ \pw{and key-frame action $a_{t_k}$ have a corresponding} step instruction $\ell_k$ describing the motion of the robot within the segment.
For example, the task "open the top drawer" is decomposed into following step instructions:
\pw{$\ell_1$:} "The robot arm lowers itself to align with the handle of the top drawer",
\pw{$\ell_2$:} "The robot arm grasps the top drawer's handle firmly", and 
\pw{$\ell_3$:} "The robot pulls the handle back, smoothly opening the top drawer".


\pw{In addition to task decomposition done at the beginning, at every execution timestep, the VLM $f_\theta$ is also exploited to predict both the step instruction $\ell_{t_k}$ \pw{(used as a novel input of the action predictor)} and its language-aligned 3D keypoint $p_{t_k}$ (for cropping a region of interest, as usually done in coarse-to-fine policies).
We discuss next how this prediction can be realized effectively.}

Task decomposition enables reasoning about task plans before predicting actions.
However, \pw{directly} training the model $f_\theta$ to \pw{simultaneously} generate a task plan $\mathcal L$ \pw{and predict step instruction $\ell_{t_k}$ and 3D keypoint $p_{t_k}$}, given \pw{as inputs} multi-view images \pw{obtained from observation $o_{t_{k-1}}$} and a high-level task description \pw{$L$}, (\pw{i.e.},  $f_{\theta}(o_{t_{k-1}}, L) \rightarrow(\mathcal{L}, \ell_{t_k}, p_{t_k})$), does not ensure generalization to novel instructions.
\pw{Previous} studies \citep{openvla, pi_05, CoTVLA, vla-os} indicate that VLAs exhibit a strong bias towards visual inputs, due to the richer information embedded in the visual inputs.
This reliance often causes failures \pw{to follow} novel language instructions.
\pw{U}nder our data-scarce training setting, this issue is intensified by the limited diversity of language instructions.
To address this issue, we propose decoupling task planning from keypoint prediction via a two-round inference protocol.
First, a purely textual query generates a language plan, the sequence of all step instructions $\mathcal{L}$.
Second, \pw{the} visual inputs \pw{augmented with this plan are used} to predict the corresponding step instruction and keypoint:
\begin{equation}\label{eq:twosteps}
    f_\theta (L_i) \rightarrow \mathcal{L}, f_\theta (o_{t_{k-1}}, \mathcal{L}) \rightarrow (\ell_{t_k}, p_{t_k}).
\end{equation}
This approach not only mitigates visual bias but also enables training with an auxiliary dataset of language plans, which serves as a manual to enhance compositional task reasoning.

\pw{For a plan $\mathcal L$ corresponding to a task description $L$, 
\pw{directly} fine-tuning a VLM to \pw{predict} both the step instruction $\ell_{t_k}$ and its language-aligned 3D keypoint $p_{t_k}$ reveals to be} insufficient \pw{due to the inadequate alignment of the visual-textual embeddings of the VLM for this 3D keypoint prediction.}
\pw{Instead,} inspired by Chain-of-Thought reasoning \citep{embodiedgpt, robotic_cot, CoTVLA} for robotics, we design a reasoning process by training our model to first reason about the pixel positions of task-related objects\pw{,} then generate a step instruction and finally predict the corresponding keypoint\pw{, changing the second step of \Cref{eq:twosteps} to:}
\begin{equation}
    f_{\theta}(o_{t_{k-1}}, \mathcal L) \rightarrow (\bm p_{obj}, \ell_{t_k}, p_{t_k}),
\end{equation}
where $\bm p_{obj}$ are the 3D positions of the task-related objects.

We further observe that for long-horizon, especially repetitive tasks (e.g., "stack several blocks"), only providing the entire task plan whether as input or output can degrade the performance.
This often causes the model to generate repetitive sequences until reaching the output length limit or to struggle in determining the next step from an excessively long plan.
To mitigate this, we introduce \pw{two ideas.
First, we provide to the VLM an additional input: the step instruction predicted in the last timestep, which serves} as a short-term memory cue to contextualize the current \pw{situation}.
Moreover, we decompose the plan into sub-tasks, \pw{i.e.,} $\mathcal{L} = \Big(\mathcal{L}_1=(\ell_1,\ell_2,...,\ell_n), \mathcal{L}_2=(\ell_{n+1},\ell_{n+2},...),... \Big)$.
For instance, "stack the blue and yellow cup on the red cup" is decomposed into two sequential sub-tasks: $\mathcal{L}_1$ "stack the blue on the red cup" followed by $\mathcal{L}_2$ "stack the yellow cup on the red cup".
The model is trained to generate only the task plan of current sub-task, preventing repetition and improving focus:
\begin{equation}
    f_\theta (L, \hat{\ell}) \rightarrow \mathcal{L}_m, f_\theta (o_{t_{k-1}}, \mathcal{L}_m, \hat{\ell}) \rightarrow (\bm p_{obj}, \ell_{t_k}, p_{t_k}),
\end{equation}
\pw{where $\hat{\ell}$ is the step instruction predicted in the last timestep and} $\mathcal{L}_m$ is the language plan of the $m^{th}$ sub-task.
At the beginning of a trajectory, where no previous timestep exists, $\hat\ell$ is defined as "the robot is currently at the initial state" to indicate the initial state.

To further enhance the generalization ability of the coarse task planner to object variations, we include an auxiliary task of predicting the object positions.
We randomly initialize diverse environments and record the RGB-D images of the scene along with the 3D positions of the objects.
Following the same pre-processing, both the RGB-D images and the 3D positions of the objects are projected into canonical views.
The projected multi-view images and pixel coordinates of the object positions in each view are used to construct an object position dataset.
This dataset is then utilized to co-train the VLM, reinforcing its spatial understanding and improving zero-shot generalization to object variations.

\subsection{Fine-grained Action Predictor}


\pw{The fine-grained action predictor uses a predicted step instruction $\ell_{t_k}$, instead of the original high-level task description $L$, enabling more precise and generalizable skill learning.}
Considering the significant domain shift between \pw{the} images \pw{focused around predicted $p_{t_{k}}$} from those used to pre-train standard VLMs, we decide to employ \pw{instead} pre-trained \pw{specialized} encoders to process these inputs.
Our feature encoding pipeline consists of three stages to construct a unified 3D-aware and language-aligned representation.
First, RGB images and step instructions are processed through vision-language encoders to establish semantic alignment between visual and textual inputs.
Second, depth images are encoded separately to extract explicit geometric structure.
Finally, we generate a 3D position embedding from pixel-wise 3D coordinates to incorporate spatial awareness.
These components are combined to form a 3D-aware, language-aligned representation for downstream fine-grained action prediction, following the architecture of prior work \citep{rvt2}.

\section{Experiments}
\label{Sec:exp}
\pw{We now} present the experimental settings and results in simulation and real-world experiments.
\subsection{Simulation Results}
\paragraph{Experimental \pw{Set-up}}
Our method is evaluated on GemBench \citep{gembench}, a benchmark specifically designed for evaluating the generalization ability of a policy.
A dataset containing 100 demonstrations per task along with a task description per trajectory is prepared for training.
This training set contains 16 tasks with 31 variations.
Within a trajectory, 4 cameras are placed at the front, left shoulder, right shoulder and wrist to collect RGB-D images as the observations.
The resolution of the original RGB-D images is 256x256 while the resolution of the projected images is 224x224.
Instead of evaluating on in-distribution tasks and variations, GemBench designs \pw{an} evaluation set \pw{containing} 4 levels of tasks\pw{, where different elements are varied:}
\begin{description}[leftmargin=0pt, itemindent=*,itemsep=-.5ex]
    \item - \textbf{Placements (L1}: \pw{same} 16 tasks (31 variations) as \pw{in} training set, but with \pw{novel} object placements.
    \item - \textbf{Rigid Objects (L2)}: 15 \pw{novel} tasks (28 variations) with \pw{newly-colored or -shaped} rigid objects.
    \item - \textbf{Articulated Objects (L3)}: 18 novel tasks (21 variations) with appearance or object variation.
    \item - \textbf{Long-horizon Tasks (L4)}: 6 \pw{novel} long-horizon tasks (12 variations).
\end{description}
The specific configuration for tasks used for training and evaluation in GemBench are listed in \Cref{appendix:gembench task specification}.
\revise{We explain in detail how we create the language plan dataset and object position dataset in \Cref{appendix:data specification}.}
Following the evaluation setting in GemBench \citep{gembench}, all trained models are evaluated with 20 episodes per task variation per seed, and 5 different seeds are used.

In our method, we finetune Qwen 2.5 VL-3B \citep{Qwen2.5-VL} as the coarse task planner.
It is LoRA fine-tuned \citep{lora} with the object keypoint dataset, language plans\pw{,} and robot trajectories.
\hjs{We use SigLIP \citep{Siglip} to extract features from the RGB images and step instructions, leveraging its language-aligned representations.
For depth images, we use DINOv2 \citep{DINOv2}, which excels at capturing geometric structures like edges and contours.
We further enhance these features by using the 3D coordinate of each pixel to construct a 3D position embedding.}
The hyperparameters, such as batch size and learning rate used in training are listed in \Cref{appendix:hyperparameters}.

To construct the fine-tuning dataset from robot trajectories, we design a sampling strategy to choose samples from the trajectories.
Apart from key-frame pairs of observation and action $\pw{(}o_{t_k}, a_{t_{k+1}}\pw{)}$, RVT2 augments the training data by sampling observations every $n$ frames (e.g., every 10 frames).
However, this results in an uneven number of samples per trajectory segment, due to \pw{the} varying length of each segment.
We initially attempt\pw{ed} to \pw{sample} observations within a window $\pw{(}o_{t_k-m},...,o_{t_k},...o_{t_k+m}\pw{)}$ around the time step $t_k$ of the $k^{th}$ key-frame.
However, observations at time steps before a key-frame and after a key-frame are visually similar while corresponding to distinct actions.
For example, $o_{t_k-1}$ and $o_{t_k+1}$ are similar while their corresponding keypoints are the gripper positions at $t_k$ and $t_{k+1}$ respectively.
Using all observations around the key-frames to fine-tune the VLM risks confusing it.
Finally, we choose to use observations $\pw{(}o_{t_k},...o_{t_k+m}\pw{)}$ at the time steps immediately following each key-frame.
We empirically choose $m$ as 5 in all experiments.

\paragraph{Main Results}
\begin{table*}[t]
\centering
\Huge
\resizebox{.9\textwidth}{!}{
\begin{tabular}{lccccccccccc} 
\toprule
\rowcolor[HTML]{CBCEFB}
Models                                          & Avg. Success $\uparrow$   & L1                 & L2                 & L3                 & L4                  \\
\midrule
HiveFormer~\citep{hiveformer}                & 30.4                                                  & 60.3 \rpmh 1.5                & 26.1 \rpmh 1.4                 & 35.1 \rpmh 1.7                 & 0.0 \rpmh 0.0           \\
\rowcolor[HTML]{EFEFEF}
PolarNet~\citep{polarnet}                   & 38.4                      & 77.7\rpmh0.9                & 37.1 \rpmh1.4                 & 38.5 \rpmh 1.7                 & 0.1 \rpmh 0.2         \\
3D Diffuser Actor~\citep{3ddiffuseractor}                       & 44.0                                          & 91.9\rpmh 0.8 & 43.4 \rpmh 2.8   & 37.0 \rpmh 2.2     & 0.0 \rpmh 0.0      \\
\rowcolor[HTML]{EFEFEF}
RVT2~\citep{rvt2}                                   & 44.0          & 89.1 \rpmh 0.8 & 51.0 \rpmh 2.3        & 36.0 \rpmh 2.2 & 0.0 \rpmh 0.0     \\
3D-LOTUS~\citep{gembench}                                 & 45.7     & \tb{94.3} \rpmh 1.4   & 49.9 \rpmh 2.2       & 38.1 \rpmh 1.1   & 0.3 \rpmh 0.3      \\
\rowcolor[HTML]{EFEFEF}
3D-LOTUS++~\citep{gembench}              & 48.0                         & 68.7 \rpmh 0.6 & 64.5 \rpmh 0.9        & 41.5 \rpmh 1.8 & 17.4 \rpmh 0.4     \\
BridgeVLA~\citep{bridgevla}                                   & 50.0                        & 91.1 \rpmh 1.1 & 65.0 \rpmh 1.3        & 43.8 \rpmh 1.2 & 0.0 \rpmh 0.0    \\
\rowcolor[HTML]{EFEFEF}
\tb{\method (Proposed)}                                   & \tb{62.0}                        & 83.9 \rpmh 0.3 & \tb{83.2} \rpmh 1.9        & \tb{49.6} \rpmh 2.1 & \tb{31.4} \rpmh 0.6    \\
\bottomrule
\end{tabular}
}
\caption{\textbf{Multi-Task Performance on GemBench.} Here are the average success rates of 4 levels of evaluation tasks from Gembench.
Except \method, we use the results reported in BridgeVLA.
}
\vspace{-4mm}
\label{table:gembench}
\end{table*}
The evaluation results on GemBench are summarized in \Cref{table:gembench}, reporting the average success rate for tasks at each generalization level.
The detailed success rate for each task are recorded in \Cref{appendix:gembench detailed results}.
The experimental results demonstrate the strong generalization ability of our method to novel tasks and object variations, as indicated by the performance gain on Level-2, Level-3, and Level-4 tasks. 
Our method achieves an overall success rate 12\% higher than prior state-of-the-art method \citep{bridgevla}.
Notably, this improvement is obtained using only 20 trajectories per task variation for training, significantly fewer than the 100 trajectories used by other baselines.
Furthermore, our design leads to substantial performance gain on the most challenging Level-4 tasks, where several baselines methods fail consistently.
\revise{We also include an experiments on the effects of using different number of trajectories in \Cref{appendix:additional ablation}.}

\paragraph{Ablation}
We further experimentally validate the design choice for both coarse task planner and fine-grained action predictor on GemBench.
The configurations are detailed below and corresponding results are presented in \Cref{table:ablation}.
\revise{A specific ablation on the inputs of the fine-grained action predictor is include in \Cref{appendix:additional ablation}.}
\begin{enumerate}[leftmargin=0pt, itemindent=*, label=\arabic*)]
    \item \tb{Base}
    In the base version (corresponding to Exp ID 1), the coarse task planner is trained with only the robot trajectories to predict step instruction and the corresponding keypoints.
    We use this version as a baseline to ablate our method.
    \item \tb{Object Reasoning}
    To adapt the pre-trained VLM for 3D keypoint prediction, we introduce a structured reasoning procedure where the model first localizes task-relevant objects before predicting the step instruction and its corresponding keypoint.
    We evaluate the efficacy of this object position reasoning in Exp ID 2.
    A comparison with the base model (Exp ID 1) reveals a performance improvement on Level-2 and Level-3 tasks, indicating enhanced generalization to object variations.

    \item \tb{Language Plan Reasoning} 
    The proposed task decomposition enables a two-round conversation, where a language plan is first generated through textual reasoning, followed by keypoint prediction.
    This approach also permits the inclusion of additional language plans during training to enhance compositional generalization.
    Compared to the base model (Exp ID 1), this version shows improvements on Level 3 and Level 4 tasks, demonstrating stronger generalization to novel task variations.
    
    \item \tb{Include Previous Step Instruction} Previous step instruction is included in the input as a short-term memory to help contextualize the current status.
    This design yields performance gains across Levels 2, 3, and 4, with particularly notable improvements on long-horizon tasks in Level 4.
    \item \tb{\method w/o Pre-trained Encoder}
    An ablation study (comparing Exp ID 5 and Exp ID 6) on the coarse planner confirms that incorporating the 3D-aware representation contributes to performance gains at all generalization levels.
    \item \tb{\method}
    Our method integrates all components mentioned above. 
    A comparison between Exp ID 4 and ours can further validate the performance gain from adding the object position dataset.

\end{enumerate}
\begin{table*}[t]
 \centering
 \setlength\tabcolsep{3pt}
 \resizebox{\textwidth}{!}{
 \begin{tabular}{cccccccccccccc}
\rowcolor[HTML]{CBCEFB}
Exp & Language Plan & Object Keypoints       & Last          &Reason & Reason    & Pretrained  & Level & Level & Level & Level & Avg.\\
\rowcolor[HTML]{CBCEFB}
ID & Data & Data        & Step      & Plan       & Objects    & Encoder    & 1 & 2 & 3 & 4 & Succ.\\
\toprule
1  & \no   & \no            & \no    &\no          & \no     &   \yes        & 86.9 & 68.2 & 36.4 & 0.4 & 48.0 \\
\rowcolor[HTML]{EFEFEF}
2  & \no    & \no            & \no   &\no           & \yes    &    \yes        & 81.8 & 74.8 & 39.0 & 0 & 48.9 \\
3  & \yes   & \no          & \no    &\yes               & \no     &      \yes      & 83.4  & 66.1  & 41.9 & 2.0 & 48.3\\
\rowcolor[HTML]{EFEFEF}
4  & \yes   & \no             & \yes  &\yes                & \yes    &    \yes      & 84.8 & 81.4 & 43.8 & 30.4 &   60.1\\
5  & \yes   & \yes             & \yes   &\yes             & \yes   &    \no       & 82.4 & 79.1 & 44.5 & 25     & 57.8\\
\rowcolor[HTML]{EFEFEF}
6  & \yes   & \yes             & \yes    &\yes          & \yes   &    \yes        & 83.8 & 83.2 & 49.6 & 31.4   & 62.0\\
\bottomrule
 \end{tabular}
 }
 ~~~~~
\caption{\textbf{Ablation study of \method on GemBench.} Here are the average success rates of 4 levels of evaluation tasks from Gembench under different training settings.}
 \vspace{-2mm}
 \label{table:ablation}
\end{table*}

\subsection{Real-world Experiments}
\begin{figure}
    \centering
    \includegraphics[width=0.99\linewidth]{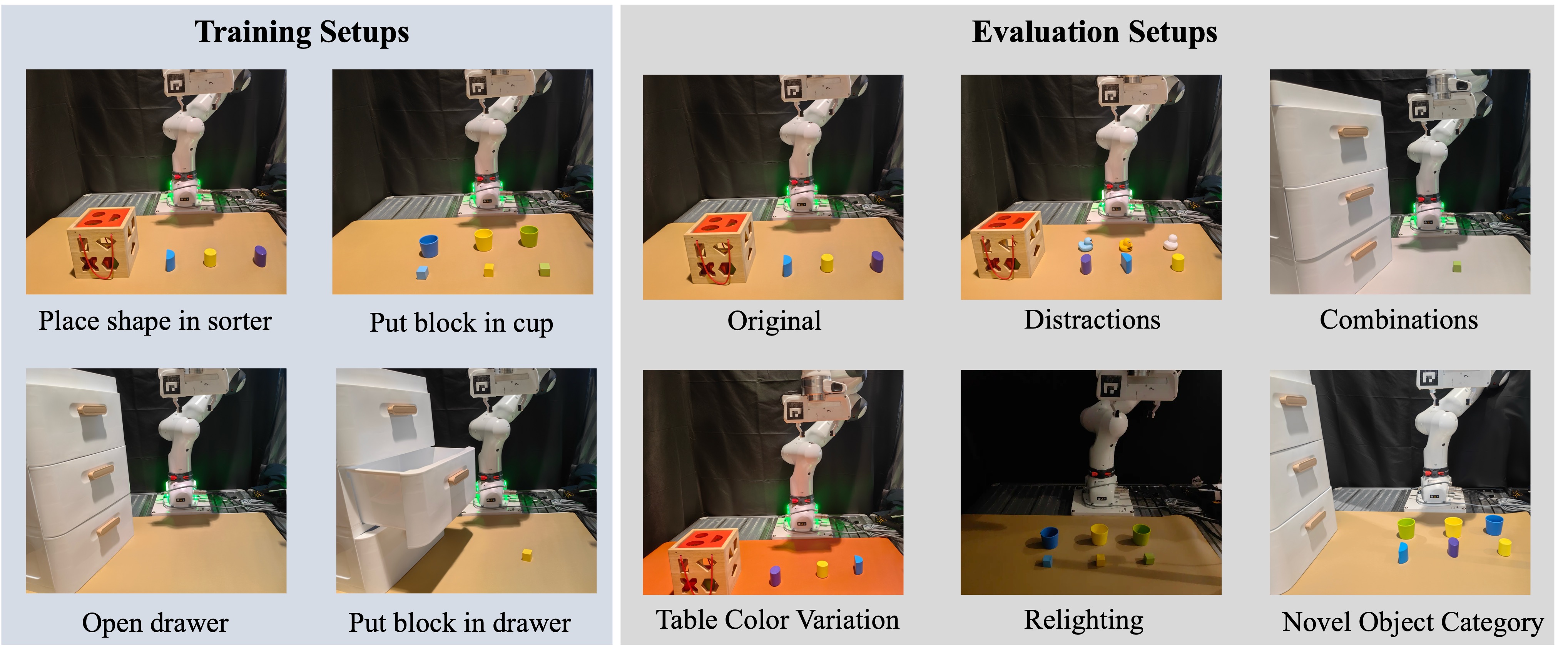}
    \caption{\tb{Overview of the tasks in real-world experiments.}
    There are four training tasks: put shape in shape sorter, put block in cup, open drawer, put block in drawer.
    We evaluate the same tasks under different visual perturbations and novel tasks designed based on the training tasks.}
    \label{fig:real_world_tasks}
\end{figure}

\paragraph{Experimental Setting}
We keep the training settings the same as in the simulation and list key modifications here.
In the real-world experiments, we use a single camera (Intel RealSense D435i) to collect RGB-D images of size 640x360.
10 trajectories are collected per task to cover all variations of each task.
The hyperparameters used for training the models are listed in \Cref{appendix:hyperparameters}.
\pw{The training tasks, illustrated in \Cref{fig:real_world_tasks}, are listed below.}
1) Place shape in shape sorter: insert objects into \pw{a} box with $3$ variations on the \pw{object} shape.
2) Put block in cup: put \pw{a} colored block in \pw{a} \pw{same-colored} cup with $3$ variations on colors.
3) Open drawer: open \pw{a} drawer with $3$ variations on the handles.
4) Put a block in drawer: put a colored block in \pw{an} open drawer with $3$ variations on color\pw{s}. 

%
%
%

We assess the generalization ability of \method along two key dimensions \pw{(see \Cref{appendix:real-world exp evaluation} for an overview of these evaluation tasks):} 
\begin{enumerate}[leftmargin=0pt, itemindent=*, label=\arabic*),itemsep=-.1ex]
    \item \tb{Visual Perturbations}:
    The model is tested on the tasks same as the training tasks but with the following conditions: different table colors, introducing distracting objects and altered backgrounds.
    \item \tb{Task/Object Variations}:
    Generalization is evaluated through: i) Object substitution (e.g., placing a "shape" object into a cup), and ii) Skill composition (e.g., combining "open drawer" and "place block" into a single, sequential task).
\end{enumerate}

\paragraph{Results}

\begin{table*}[t]
\centering
\Huge
\resizebox{\textwidth}{!}{
\begin{tabular}{lccccccccccc} 
\toprule
\rowcolor[HTML]{CBCEFB}
 & \multicolumn{2}{c}{No Variation} & \multicolumn{2}{c}{Table Color} & \multicolumn{2}{c}{Distracted Objects} & \multicolumn{2}{c}{Light Strength} & \multicolumn{2}{c}{Average Succ.}\\
\cmidrule(lr){2-3} \cmidrule(lr){4-5} \cmidrule(lr){6-7} \cmidrule(lr){8-9} \cmidrule(lr){10-11}
\rowcolor[HTML]{CBCEFB}
 & RVT2 & \method & RVT2 & \method & RVT2 & \method & RVT2 & \method & RVT2 & \method \\
\midrule
place shape in shape sorter   &  60\% &  \tb{60\%} & 35\% & \tb{50\%}  &  30\% &  \tb{40\%} & 20\% & \tb{50\%}   & 36.2\% & \tb{50\%}     \\
\rowcolor[HTML]{EFEFEF}
put block in cup with same color & 40\% &  \tb{100\%} &  20\% &\tb{70\%} &  20\% &  \tb{80\%} & 20\%  & \tb{80\%}  & 25\% & \tb{82.5\%}  \\
open drawer                      & 100\% & \tb{100\%} & 85\% & \tb{95\%}  &  100\% &  \tb{100\%} & 0\%  & \tb{100\%}  & 71.2\% & \tb{98.7\%}   \\
\rowcolor[HTML]{EFEFEF}
put block in open drawer         & 40\%  &  \tb{90\%} & 25\% & \tb{90\%} & 20\% &  \tb{70\%} &  0\%  & \tb{90\%}  & 21.2\% & \tb{85\%}  \\
\midrule
put block in cup with different color    &  20\% &  \tb{100\%} &  15\% & \tb{70\%} & 0\% &  \tb{70\%} & 10\%   & \tb{80\%}   & 11.2\% & \tb{80\%}    \\
\rowcolor[HTML]{EFEFEF}
put shape in open drawer         &  30\% &  \tb{80\%} &  20\% & \tb{65\%} & 10\% &  \tb{70\%}  &  0\%   & \tb{80\%}  & 15\% & \tb{73.7\%}   \\
put shape in cup                 &  0\%  &  \tb{80\%} & 0\% & \tb{70\%} & 0\% &  \tb{60\%} & 0\%   & \tb{70\%}  & 0\% & \tb{70\%}    \\
\rowcolor[HTML]{EFEFEF}
put block in close drawer       &  0\% &  \tb{90\%} &  0\% & \tb{75\%} &  0\% &  \tb{70\%} & 0\%  &  \tb{80\%}   & 0\%  & \tb{78.7\%}    \\
\midrule
average  success rate    & 36.2\%  & \tb{87.5\%} & 25\%  & \tb{73.1\%} & 22.5\%  & \tb{70\%}  & 6.2\%  &  \tb{78.7\%}     & 22.5\% & \tb{77.3\%} \\
\bottomrule
\end{tabular}
}
\caption{\textbf{Real-world Performance.} Here are the average success rate under different generalization settings for real-world experiments.
}
\vspace{-4mm}
\label{table:real_world}
\end{table*}

The results of evaluating the trained models in real-world experiments are summarized in \Cref{table:real_world}.
Our method achieves a strong generalization ability to novel tasks and object variations, trained with only 10 demonstrations per task.
\method achieves 54.8\% higher average success rates compared to RVT2 on the evaluation tasks.

\section{Conclusions}
We propose a novel coarse-to-fine 3D manipulation policy, where tasks are decomposed and pre-trained models are leveraged in the hierarchical architecture.
Our method demonstrates strong generalization capabilities while maintaining the sample efficiency inherent to coarse-to-fine approaches.
Although leveraging pre-trained models for robotics tasks is common, their effective adaptation for generalizable and precise is still under-explored.
We hope this work inspires further research into building highly generalizable and sample-efficient 3D manipulation policies.
Our method has two key limitations.
First, key-frame based imitation learning is suitable for structured tasks that can be easily decomposed into discrete steps.
Unstructured tasks, such as wiping a desk, where key-frames are difficult to define, present a significant challenge.
Moreover, the current framework lacks a robust error-correction mechanism.
An incorrect action prediction at any step might lead to task failure.
A promising future direction is to integrate a self-correction module to enhance robustness.

\paragraph{Acknowledgement}
This work has been supported in part by the program of National Natural Science Foundation of China (No. 62176154), the program of National Natural Science Foundation
of China (No. 62503322), Shanghai Magnolia Funding Pujiang Program (No.
23PJ1404400), and the AI for Science Seed Program of Shanghai Jiao Tong University (project number 2025AI4S-QY06).

\paragraph{LLM Usage}
We used Deepseek \citep{deepseekv3} and ChatGPT \citep{chatgpt4o} for grammar check and related work retrieval.
The authors have reviewed the content generated by the LLM.

\paragraph{Ethics Statement}
We adhere to the ICLR Code of Ethics and take full responsibility for the final content.

\paragraph{Reproducibility Statement}
To ensure reproducibility, we provide a comprehensive description of our method and experimental setup in the \Cref{Sec:method} and the \Cref{Sec:exp}, document all hyperparameters in the appendix \Cref{tab:hyper}, and release our code at \url{https://github.com/Jianshu-Hu/Generalizable-CLAP}.

\bibliography{iclr2026_conference}
\bibliographystyle{iclr2026_conference}

\appendix
\clearpage
\section{Appendix}
\subsection{GemBench Task Specification}
\label{appendix:gembench task specification}
The detailed tasks and variations in GemBench \citep{gembench} used for training and evaluation are listed in \Cref{tab:gembench_tasks}.

\revise{\subsection{Training Data Specification}
\label{appendix:data specification}
We have three datasets in total for finetuning the pre-trained VLM: a language plan dataset, a object position dataset and a robot trajectory dataset.
\begin{itemize}
    \item For language plan dataset, we label each task with a language plan according to the decomposition of the task.
    Note that one task may contain different variations.
    Each variation of the task is assigned with a different language plan.
    We augment the data by asking LLM to provide some variations on the language descriptions of the plan. 
    \item For object position dataset, we randomly initialize the scene of the tasks and apply transformations (translation and rotation) on the scene.
    We record the RGB-D images along with the 3D positions of the objects.
    The 3D positions of the objects are projected into canonical views to obtain the pixel positions in each view.
    \item For robot trajectory dataset, we explain in the Section 5.1 on how we sample transitions from the demonstrations to build observation-action pair.
    Transformations (translation and rotation) are also applied on the transitions to augment the data.
\end{itemize}
In practice, the number of training samples for each dataset used for GemBench is listed in \Cref{tab:number_of_samples}.
}

\begin{table}[tb]
\centering
\caption{\textbf{Number of samples.}
We record the number of samples in training set and validation set for different datasets used in the simulation experiments for GemBench.}
\label{tab:number_of_samples}
\begin{tabular}{lccc}
           & \multicolumn{1}{l}{Language Plan} & \multicolumn{1}{l}{Object Position} & \multicolumn{1}{l} {Robot Trajectory Data} \\ \hline
Train      & 30515                             & 39777                               & 169665                \\
Validation & 6103                              & 16650                               & 17400                  \\ \hline
\end{tabular}
\end{table}

\begin{table*}[t]
\centering
\tabcolsep=0.18cm
\caption{\textbf{Training and evaluation tasks \& variations in GemBench.}
The evaluation tasks contain four levels of generalization, where Level 1 evaluates the generalization to novel placements, Level 2 novel rigid objects, Level 3 novel articulated objects, and Level 4 novel long-horizon tasks.}
\label{tab:gembench_tasks}
\resizebox{\textwidth}{!}{
\begin{tabular}{llccccccc}
\toprule
\rowcolor[HTML]{CBCEFB} 
\cellcolor[HTML]{CBCEFB} & \multicolumn{2}{c}{\cellcolor[HTML]{CBCEFB}Train / Level 1} & \multicolumn{2}{c}{\cellcolor[HTML]{CBCEFB}Level 2} & \multicolumn{3}{c}{\cellcolor[HTML]{CBCEFB}Level 3} & Level 4 \\
\rowcolor[HTML]{CBCEFB} 
\multirow{-2}{*}{\cellcolor[HTML]{CBCEFB}} & Task & Variation & Color & Shape & Instance & Category & Action-Part & Long-horizon \\ \midrule
 &  & maroon button & azure button &  &  &  &  & 2 buttons \\
 &  & navy button & rose button &  &  &  &  & 3 buttons \\
\multirow{-3}{*}{Press} & \multirow{-3}{*}{Push button} & yellow button & white button & \multirow{-3}{*}{Lamp on} &  &  &  & 4 buttons \\ \midrule
\rowcolor[HTML]{EFEFEF} 
\cellcolor[HTML]{EFEFEF} & \cellcolor[HTML]{EFEFEF} & red block & teal block & red cylinder &  &  &  &  \\
\rowcolor[HTML]{EFEFEF} 
\cellcolor[HTML]{EFEFEF} & \cellcolor[HTML]{EFEFEF} & lime block & violet block & red star &  &  &  &  \\
\rowcolor[HTML]{EFEFEF} 
\cellcolor[HTML]{EFEFEF} & \multirow{-3}{*}{\cellcolor[HTML]{EFEFEF}Pick and lift} & cyan block & black block & red moon &  &  &  &  \\ %
\rowcolor[HTML]{EFEFEF} 
\cellcolor[HTML]{EFEFEF} & \cellcolor[HTML]{EFEFEF} & magenta cup & gray cup & \cellcolor[HTML]{EFEFEF} &  &  &  &  \\
\rowcolor[HTML]{EFEFEF} 
\cellcolor[HTML]{EFEFEF} & \cellcolor[HTML]{EFEFEF} & silver cup & olive cup & \cellcolor[HTML]{EFEFEF} &  &  &  &  \\
\rowcolor[HTML]{EFEFEF} 
\multirow{-6}{*}{\cellcolor[HTML]{EFEFEF}Pick} & \multirow{-3}{*}{\cellcolor[HTML]{EFEFEF}Pick up cup} & orange cup & purple cup & \multirow{-3}{*}{\cellcolor[HTML]{EFEFEF}red toy} &  &  &  &  \\ \midrule
 &  & green target & pink target &  &  &  &  &  \\
 & \multirow{-2}{*}{Slide block} & blue target & yellow target &  &  &  &  &  \\ %
 &  & teal target & cyan target &  &  &  &  &  \\ 
\multirow{-4}{*}{Push} & \multirow{-2}{*}{Reach and drag} & black target & navy target &  &  &  &  &  \\ \midrule
\rowcolor[HTML]{EFEFEF} 
\cellcolor[HTML]{EFEFEF} & \cellcolor[HTML]{EFEFEF} & azure jar & blue jar &  &  &  &  &  \\
\rowcolor[HTML]{EFEFEF} 
\cellcolor[HTML]{EFEFEF} & \multirow{-2}{*}{\cellcolor[HTML]{EFEFEF}Close jar} & violet jar & green jar &  &  &  &  &  \\ %
\rowcolor[HTML]{EFEFEF} 
\cellcolor[HTML]{EFEFEF} & \cellcolor[HTML]{EFEFEF} & rose bulb & lime bulb &  &  &  &  &  \\
\rowcolor[HTML]{EFEFEF} 
\multirow{-4}{*}{\cellcolor[HTML]{EFEFEF}Screw} & \multirow{-2}{*}{\cellcolor[HTML]{EFEFEF}Screw bulb} & white bulb & maroon bulb &  &  &  &  &  \\ \midrule
 & Close fridge & fridge &  &  & fridge2 &  & door &  \\
 & Close laptop lid & laptop lid &  &  & laptop lid2 &  & box &  \\
\multirow{-3}{*}{Close} & Close microwave & microwave &  &  & microwave2 & \multirow{-3}{*}{grill} & drawer &  \\ \midrule
\rowcolor[HTML]{EFEFEF} 
\cellcolor[HTML]{EFEFEF} & Open door & door &  &  & door2 & \cellcolor[HTML]{EFEFEF} & fridge & \cellcolor[HTML]{EFEFEF} \\
\rowcolor[HTML]{EFEFEF} 
\cellcolor[HTML]{EFEFEF} & Open box & box &  &  & box2 & \multirow{-1}{*}{\cellcolor[HTML]{EFEFEF}toilet} & laptop lid & \multirow{-2}{*}{\cellcolor[HTML]{EFEFEF}\begin{tabular}[c]{@{}c@{}}Take shoes \\ out of box\end{tabular}} \\
\rowcolor[HTML]{EFEFEF} 
\cellcolor[HTML]{EFEFEF} & \cellcolor[HTML]{EFEFEF} & bottom drawer &  &  & drawer2, drawer3 & \cellcolor[HTML]{EFEFEF} & microwave & \cellcolor[HTML]{EFEFEF} \\
\rowcolor[HTML]{EFEFEF} 
\multirow{-4}{*}{\cellcolor[HTML]{EFEFEF}Open} & \multirow{-2}{*}{\cellcolor[HTML]{EFEFEF}Open drawer} & top drawer &  &  & long drawer w/ 4 levels &  & middle drawer & \multirow{-2}{*}{\cellcolor[HTML]{EFEFEF}\begin{tabular}[c]{@{}c@{}}Put 3 items \\ in drawer\end{tabular}} \\ \midrule
 &  & 2 gray blocks & 2 orange blocks &  &  &  &  &  \\
 &  & 2 olive blocks & 2 silver blocks &  &  &  &  & \multirow{-2}{*}{\begin{tabular}[c]{@{}c@{}}Stack 3-4 \\ blocks\end{tabular}} \\
 & \multirow{-3}{*}{Stack blocks} & 2 purple blocks & 2 magenta blocks &  &  &  &  & Stack 2 cups \\ \cmidrule{2-9}
 &  & crackers box &  & mustard bottle &  &  &  &  \\
 & \multirow{-2}{*}{Put groceries} & soup can &  & sugar box &  &  &  & \multirow{-2}{*}{\begin{tabular}[c]{@{}c@{}}Put all \\ groceries\end{tabular}} \\ \cmidrule{2-9} 
 &  & bottom shelf &  &  &  &  &  &  \\
\multirow{-7}{*}{\begin{tabular}[c]{@{}c@{}}Put/\\ Stack\end{tabular}} & \multirow{-2}{*}{Put money} & middle shelf &  & \multirow{-2}{*}{\begin{tabular}[c]{@{}c@{}}Put cube in \\ bottom shelf\end{tabular}} &  &  & \multirow{-2}{*}{top shelf} &  \\ \bottomrule
\end{tabular}
}
\end{table*}

\subsection{Experimental details}
\label{appendix:hyperparameters}
All experiments are conducted on 4 NVIDIA RTX 4090 GPU.
The hyperparameters and training time are listed in \Cref{tab:hyper}.

\begin{table}[]
\caption{\tb{Training time and hyperparameters used in different experiments.} Here we list the training time and hyperparameters used for training the model with GemBench and real-world data.}
\label{tab:hyper}
\resizebox{\textwidth}{!}{
\begin{tabular}{lcccc}
\cline{2-5}
               & \multicolumn{2}{c}{GemBench}                        & \multicolumn{2}{c}{Real-world}                      \\ \cline{2-5} 
               & Coarse task planner & Fine-grained Action Predictor & Coarse task planner & Fine-grained Action Predictor \\ \hline
Training time  & 6 hours             & 3 hours                       & 1 hour              & 1 hour                        \\
Learning rata  & 3e-4                & 0.0024                        & 1e-4                & 0.0024                        \\
Batch Size     & 64                  & 192                           & 64                  & 192                           \\
Epochs         & 1                   & 5                             & 1                   & 3                             \\
Lora Rank      & 8                   & /\                             & 8                   & /\                             \\
Lora Alpha     & 32                  & /\                             & 32                  & /\                             \\
Freeze Vit     & False               & /\                             & False               & /\                             \\
Freeze Aligner & True                & /\                             & True                & /\                             \\
Freeze LLM     & False               & /\                             & False               & /\                             \\ \hline
\end{tabular}
}
\end{table}

\subsection{GemBench Success Rate Per Task}
\label{appendix:gembench detailed results}
\begin{table*}[t]
\centering
\resizebox{\textwidth}{!}{ 
\begin{tabular}{lccccccccccc} \toprule
\rowcolor[HTML]{CBCEFB}
Method & Avg. & \begin{tabular}[c]{@{}c@{}}Close\\ Fridge+0\end{tabular} & \begin{tabular}[c]{@{}c@{}}Close\\ Jar+15\end{tabular} & \begin{tabular}[c]{@{}c@{}}Close\\ Jar+16\end{tabular} & \begin{tabular}[c]{@{}c@{}}CloseLaptop\\ Lid+0\end{tabular} & \begin{tabular}[c]{@{}c@{}}Close\\ Microwave+0\end{tabular} & \begin{tabular}[c]{@{}c@{}}LightBulb\\ In+17\end{tabular} & \begin{tabular}[c]{@{}c@{}}LightBulb\\ In+19\end{tabular} & \begin{tabular}[c]{@{}c@{}}Open\\ Box+0\end{tabular} & \begin{tabular}[c]{@{}c@{}}Open\\ Door+0\end{tabular} & \begin{tabular}[c]{@{}c@{}}Open\\ Drawer+0\end{tabular} \\ \midrule
HiveFormer~\citep{hiveformer} & 60.3$_{\pm 1.5}$ & 96$_{\pm 4.2}$ & 64$_{\pm 13.9}$ & 92$_{\pm 2.7}$ & 90$_{\pm 3.5}$ & 88$_{\pm 7.6}$ & 12$_{\pm 4.5}$ & 13$_{\pm 6.7}$ & 4$_{\pm 4.2}$ & 53$_{\pm 15.2}$ & 15$_{\pm 12.2}$ \\
\rowcolor[HTML]{EFEFEF}
PolarNet~\citep{polarnet} & 77.6$_{\pm 0.9}$ & 99$_{\pm 2.2}$ & 99$_{\pm 2.2}$ & 99$_{\pm 2.2}$ & 95$_{\pm 3.5}$ & 98$_{\pm 2.7}$ & 72$_{\pm 12.5}$ & 71$_{\pm 6.5}$ & 32$_{\pm 11.5}$ & 69$_{\pm 8.9}$ & 61$_{\pm 12.4}$ \\
3D Diffuser Actor~\citep{3ddiffuseractor}  & 91.9$_{\pm 0.8}$ & \textbf{100}$_{\pm 0.0}$ & \textbf{100}$_{\pm 0.0}$ & \textbf{100}$_{\pm 0.0}$ & \textbf{99}$_{\pm 2.2}$ & \textbf{100}$_{\pm 0.0}$ & 85$_{\pm 5.0}$ & 88$_{\pm 2.7}$ & 11$_{\pm 2.2}$ & 96$_{\pm 4.2}$ & 82$_{\pm 9.1}$ \\
\rowcolor[HTML]{EFEFEF}
RVT2~\citep{rvt2}  & 89.0$_{\pm 0.8}$ & 77$_{\pm 11.0}$ & 97$_{\pm 4.5}$ & 98$_{\pm 2.7}$ & 77$_{\pm 13.0}$ & \textbf{100}$_{\pm 0.0}$ & \textbf{93}$_{\pm 5.7}$ & \textbf{91}$_{\pm 8.2}$ & 7$_{\pm 4.5}$ & \textbf{98}$_{\pm 4.5}$ & \textbf{93}$_{\pm 5.7}$ \\
3D-LOTUS~\citep{gembench} & \textbf{94.3}$_{\pm 3.5}$ & 96$_{\pm 3.7}$ & \textbf{100}$_{\pm 0.0}$ & \textbf{100}$_{\pm 0.0}$ & 98$_{\pm 2.5}$ & 98$_{\pm 4.0}$ & 84$_{\pm 7.4}$ & 85$_{\pm 9.5}$ & \textbf{99}$_{\pm 2.0}$ & 77$_{\pm 2.5}$ & 83$_{\pm 8.7}$ \\ 
\rowcolor[HTML]{EFEFEF}
3D-LOTUS++~\citep{gembench} & 68.7$_{\pm 0.6}$ & 95$_{\pm 0.0}$ & \textbf{100$_{\pm 0.0}$} & 99$_{\pm 2.0}$ & 28$_{\pm 2.5}$ & 87$_{\pm 5.1}$ & 55$_{\pm 10.5}$ & 45$_{\pm 8.9}$ & 55$_{\pm 8.9}$ & 79$_{\pm 9.7}$ & 68$_{\pm 12.5}$ \\
BridgeVLA~\citep{bridgevla} & 91.1$_{\pm 1.1}$ & 99$_{\pm 2.0}$ & 98$_{\pm 4.0}$ & \textbf{100}$_{\pm 0.0}$ & 97$_{\pm 2.5}$ & 85$_{\pm 5.5}$ & 90$_{\pm 5.5}$ & 87$_{\pm 7.5}$ & 76$_{\pm 10.2}$ & 70$_{\pm 12.3}$ & 86$_{\pm 5.8}$ \\
\rowcolor[HTML]{EFEFEF}
\method & $83.9_{\pm 0.3}$ & 88$_{\pm 4.5}$ & 98$_{\pm 2.7}$ & \textbf{100}$_{\pm 0.0}$ & 88$_{\pm 9.1}$ & 99$_{\pm 2.2}$ & $84_{\pm 6.5}$ & 76$_{\pm 9.1}$ & 17$_{\pm 5.7}$ & 82$_{\pm 4.5}$ & 87$_{\pm 13.0}$ \\
\midrule

\rowcolor[HTML]{CBCEFB}
Method & \begin{tabular}[c]{@{}c@{}}Open\\ Drawer+2\end{tabular} & \begin{tabular}[c]{@{}c@{}}Pick\&\\ Lift+0\end{tabular} & \begin{tabular}[c]{@{}c@{}}Pick\&\\ Lift+2\end{tabular} & \begin{tabular}[c]{@{}c@{}}Pick\&\\ Lift+7\end{tabular} & \begin{tabular}[c]{@{}c@{}}PickUp\\ Cup+8\end{tabular} & \begin{tabular}[c]{@{}c@{}}PickUp\\ Cup+9\end{tabular} & \begin{tabular}[c]{@{}c@{}}PickUp\\ Cup+11\end{tabular} & \begin{tabular}[c]{@{}c@{}}Push\\ Button+0\end{tabular} & \begin{tabular}[c]{@{}c@{}}Push\\ Button+3\end{tabular} & \begin{tabular}[c]{@{}c@{}}Push\\ Button+4\end{tabular} & \begin{tabular}[c]{@{}c@{}}PutIn\\ Cupboard+0\end{tabular} \\ \midrule
HiveFormer~\citep{hiveformer}  & 59$_{\pm 7.4}$ & 86$_{\pm 4.2}$ & 92$_{\pm 6.7}$ & 93$_{\pm 2.7}$ & 83$_{\pm 7.6}$ & 69$_{\pm 12.9}$ & 61$_{\pm 19.8}$ & 84$_{\pm 11.9}$ & 68$_{\pm 6.7}$ & 87$_{\pm 7.6}$ & 34$_{\pm 8.2}$ \\
\rowcolor[HTML]{EFEFEF}
PolarNet~\citep{polarnet} & 90$_{\pm 7.1}$ & 92$_{\pm 9.1}$ & 84$_{\pm 7.4}$ & 88$_{\pm 5.7}$ & 82$_{\pm 7.6}$ & 79$_{\pm 4.2}$ & 72$_{\pm 10.4}$ & \textbf{100}$_{\pm 0.0}$ & \textbf{100}$_{\pm 0.0}$ & 99$_{\pm 2.2}$ & 52$_{\pm 7.6}$ \\
3D Diffuser Actor~\citep{3ddiffuseractor}  & 97$_{\pm 4.5}$ & \textbf{99}$_{\pm 2.2}$ & 99$_{\pm 2.2}$ & 99$_{\pm 2.2}$ & 96$_{\pm 2.2}$ & 97$_{\pm 4.5}$ & 98$_{\pm 2.7}$ & 98$_{\pm 2.7}$ & 96$_{\pm 4.2}$ & 98$_{\pm 2.7}$ & 85$_{\pm 5.0}$ \\
\rowcolor[HTML]{EFEFEF}
RVT2~\citep{rvt2}  & 94$_{\pm 4.2}$ & \textbf{99}$_{\pm 2.2}$ & 98$_{\pm 2.7}$ & \textbf{100}$_{\pm 0.0}$ & \textbf{99}$_{\pm 2.2}$ & \textbf{99}$_{\pm 2.2}$ & \textbf{99}$_{\pm 2.2}$ & \textbf{100}$_{\pm 0.0}$ & \textbf{100}$_{\pm 0.0}$ & \textbf{100}$_{\pm 0.0}$ & 88$_{\pm 8.4}$ \\
3D-LOTUS~\citep{gembench} & 93$_{\pm 6.0}$ & \textbf{99}$_{\pm 2.0}$ & \textbf{100}$_{\pm 0.0}$ & 99$_{\pm 2.0}$ & 97$_{\pm 4.0}$ & 96$_{\pm 3.7}$ & 94$_{\pm 4.9}$ & 99$_{\pm 2.0}$ & 99$_{\pm 2.0}$ & \textbf{100}$_{\pm 0.0}$ & \textbf{89}$_{\pm 5.8}$ \\ 
\rowcolor[HTML]{EFEFEF}
3D-LOTUS++~\citep{gembench} & 75$_{\pm 4.5}$ & 97$_{\pm 6.0}$ & 94$_{\pm 3.7}$ & 93$_{\pm 5.1}$ & 86$_{\pm 8.0}$ & 88$_{\pm 6.8}$ & 91$_{\pm 4.9}$ & \textbf{100$_{\pm 0.0}$} & \textbf{100$_{\pm 0.0}$} & \textbf{100$_{\pm 0.0}$} & 1$_{\pm 2.0}$ \\
BridgeVLA~\citep{bridgevla} & \textbf{99}$_{\pm 2.0}$ & \textbf{99}$_{\pm 2.0}$ & \textbf{100}$_{\pm 0.0}$ & 98$_{\pm 2.5}$ & 96$_{\pm 2.0}$ & 94$_{\pm 3.7}$ & \textbf{99}$_{\pm 2.0}$ & \textbf{100}$_{\pm 0.0}$ & 98$_{\pm 4.0}$ & 98$_{\pm 4.0}$ & 74$_{\pm 6.6}$ \\
\rowcolor[HTML]{EFEFEF}
\method & 98$_{\pm 2.7}$ & 98$_{\pm 2.7}$ & 99$_{\pm 2.2}$ & 99$_{\pm 2.2}$ & 94$_{\pm 5.5}$ & 99$_{\pm 2.2}$ & 93$_{\pm 7.6}$ & \textbf{100}$_{\pm 0.0}$ & \tb{100}$_{\pm 0.0}$ & 97$_{\pm 2.7}$ & 58$_{\pm 10.4}$ \\
\midrule

\rowcolor[HTML]{CBCEFB}
Method & \begin{tabular}[c]{@{}c@{}}PutIn\\ Cupboard+3\end{tabular} & \begin{tabular}[c]{@{}c@{}}PutMoney\\ InSafe+0\end{tabular} & \begin{tabular}[c]{@{}c@{}}PutMoney\\ InSafe+1\end{tabular} & \begin{tabular}[c]{@{}c@{}}Reach\&\\ Drag+14\end{tabular} & \begin{tabular}[c]{@{}c@{}}Reach\&\\ Drag+18\end{tabular} & \begin{tabular}[c]{@{}c@{}}Slide\\ Block+0\end{tabular} & \begin{tabular}[c]{@{}c@{}}Slide\\ Block+1\end{tabular} & \begin{tabular}[c]{@{}c@{}}Stack\\ Blocks+30\end{tabular} & \begin{tabular}[c]{@{}c@{}}Stack\\ Blocks+36\end{tabular} & \begin{tabular}[c]{@{}c@{}}Stack\\ Blocks+39\end{tabular} &  \\ \midrule
HiveFormer~\citep{hiveformer}  & 74$_{\pm 6.5}$ & 85$_{\pm 3.5}$ & 88$_{\pm 2.7}$ & 37$_{\pm 5.7}$ & 32$_{\pm 7.6}$ & 99$_{\pm 2.2}$ & 91$_{\pm 12.4}$ & 6$_{\pm 5.5}$ & 7$_{\pm 4.5}$ & 6$_{\pm 4.2}$ &  \\
\rowcolor[HTML]{EFEFEF}
PolarNet~\citep{polarnet}  & \textbf{88}$_{\pm 4.5}$ & 93$_{\pm 4.5}$ & 95$_{\pm 5.0}$ & 99$_{\pm 2.2}$ & 99$_{\pm 2.2}$ & \textbf{100}$_{\pm 0.0}$ & 0$_{\pm 0.0}$ & 34$_{\pm 10.8}$ & 30$_{\pm 9.4}$ & 36$_{\pm 12.9}$ &  \\
3D Diffuser Actor~\citep{3ddiffuseractor}  & 82$_{\pm 11.5}$ & \textbf{95}$_{\pm 5.0}$ & 98$_{\pm 2.7}$ & \textbf{100}$_{\pm 0.0}$ & 99$_{\pm 2.2}$ & \textbf{100}$_{\pm 0.0}$ & 89$_{\pm 4.2}$ & 88$_{\pm 7.6}$ & 85$_{\pm 6.1}$ & 89$_{\pm 5.5}$ &  \\
\rowcolor[HTML]{EFEFEF}
RVT2~\citep{rvt2}  & 80$_{\pm 6.1}$ & 93$_{\pm 8.4}$ & 96$_{\pm 8.5}$ & 85$_{\pm 10.0}$ & 94$_{\pm 2.2}$ & \textbf{100}$_{\pm 0.0}$ & 37$_{\pm 6.7}$ & 88$_{\pm 5.7}$ & \textbf{93}$_{\pm 2.7}$ & 88$_{\pm 11.5}$ &  \\
3D-LOTUS~\citep{gembench} & 72$_{\pm 11.2}$ & 94$_{\pm 3.7}$ & \textbf{99}$_{\pm 2.0}$ & 99$_{\pm 2.0}$ & \textbf{100}$_{\pm 0.0}$ & \textbf{100}$_{\pm 0.0}$ & \textbf{100}$_{\pm 0.0}$ & 94$_{\pm 5.8}$ & 91$_{\pm 6.6}$ & \textbf{90}$_{\pm 4.5}$ & \\ 
\rowcolor[HTML]{EFEFEF}
3D-LOTUS++~\citep{gembench} & 2$_{\pm 2.5}$ & 22$_{\pm 6.8}$ & 16$_{\pm 4.9}$ & 94$_{\pm 3.7}$ & 62$_{\pm 8.7}$ & \textbf{100$_{\pm 0.0}$} & 65$_{\pm 5.5}$ & 86$_{\pm 5.8}$ & 20$_{\pm 4.5}$ & 28$_{\pm 13.6}$ & \\
BridgeVLA~\citep{bridgevla} & 84$_{\pm 6.6}$ & 79$_{\pm 9.7}$ & 86$_{\pm 3.7}$ & 96$_{\pm 5.8}$ & 97$_{\pm 4.0}$ & \textbf{100}$_{\pm 0.0}$ & 90$_{\pm 5.5}$ & 77$_{\pm 8.1}$ & 87$_{\pm 4.0}$ & 85$_{\pm 7.8}$ & \\
\rowcolor[HTML]{EFEFEF}
\method & 69$_{\pm 12.4}$ & 80$_{\pm 6.1}$ & 82$_{\pm 7.6}$ & 90$_{\pm 3.5}$ & 90$_{\pm 34.5}$ & 55$_{\pm 5.0}$ & 5$_{\pm 5.0}$ & \tb{96}$_{\pm 4.2}$ & 85$_{\pm 3.5}$ & 90$_{\pm 6.1}$ & \\
\bottomrule
\end{tabular}
}
\caption{\textbf{Per-task Success Rate on GemBench Level 1.}}
\label{tab:gembench_l1_all}
\end{table*}

\begin{table*}[t]
\centering
\resizebox{\textwidth}{!}{ 
\begin{tabular}{lcccccccccc} \toprule
\rowcolor[HTML]{CBCEFB}
Method & Avg. & \begin{tabular}[c]{@{}c@{}}Push\\ Button+13\end{tabular} & \begin{tabular}[c]{@{}c@{}}Push\\ Button+15\end{tabular} & \begin{tabular}[c]{@{}c@{}}Push\\ Button+17\end{tabular} & \begin{tabular}[c]{@{}c@{}}Pick\&\\ Lift+14\end{tabular} & \begin{tabular}[c]{@{}c@{}}Pick\&\\ Lift+16\end{tabular} & \begin{tabular}[c]{@{}c@{}}Pick\&\\ Lift+18\end{tabular} & \begin{tabular}[c]{@{}c@{}}PickUp\\ Cup+10\end{tabular} & \begin{tabular}[c]{@{}c@{}}PickUp\\ Cup+12\end{tabular} & \begin{tabular}[c]{@{}c@{}}PickUp\\ Cup+13\end{tabular} \\ \midrule
HiveFormer~\citep{hiveformer} & 26.1$_{\pm 1.4}$ & 97$_{\pm 2.7}$ & 85$_{\pm 10.0}$ & 88$_{\pm 2.7}$ & 21$_{\pm 6.5}$ & 9$_{\pm 4.2}$ & 8$_{\pm 6.7}$ & 30$_{\pm 7.1}$ & 22$_{\pm 13.5}$ & 26$_{\pm 10.6}$ \\
\rowcolor[HTML]{EFEFEF}
PolarNet~\citep{polarnet} & 37.1$_{\pm 1.4}$ & 100$_{\pm 0.0}$ & \textbf{100$_{\pm 0.0}$} & 85$_{\pm 7.9}$ & 3$_{\pm 4.5}$ & 1$_{\pm 2.2}$ & 0$_{\pm 0.0}$ & 48$_{\pm 11.0}$ & 46$_{\pm 8.9}$ & 16$_{\pm 6.5}$ \\
3D Diffuser Actor~\citep{3ddiffuseractor} & 43.4$_{\pm 2.8}$ & 87$_{\pm 13.0}$ & 81$_{\pm 6.5}$ & 60$_{\pm 9.4}$ & 9$_{\pm 4.2}$ & 18$_{\pm 9.1}$ & 0$_{\pm 0.0}$ & 84$_{\pm 5.5}$ & 60$_{\pm 11.7}$ & 62$_{\pm 13.0}$ \\
\rowcolor[HTML]{EFEFEF}
RVT2~\citep{rvt2} & 51.0$_{\pm 2.3}$ & \textbf{100$_{\pm 0.0}$} & \textbf{100$_{\pm 0.0}$} & \textbf{100$_{\pm 0.0}$} & 47$_{\pm 7.6}$ & 29$_{\pm 9.6}$ & 8$_{\pm 4.5}$ & 81$_{\pm 8.2}$ & 59$_{\pm 9.6}$ & 72$_{\pm 9.7}$ \\
3D-LOTUS~\citep{gembench} & 49.9$_{\pm 2.2}$ & 99$_{\pm 2.0}$ & \textbf{100$_{\pm 0.0}$} & \textbf{100$_{\pm 0.0}$} & 3$_{\pm 2.5}$ & 18$_{\pm 8.7}$ & 33$_{\pm 9.3}$ & 89$_{\pm 3.7}$ & 78$_{\pm 8.7}$ & 57$_{\pm 7.5}$ \\ 
\rowcolor[HTML]{EFEFEF}
3D-LOTUS++~\citep{gembench} & 64.5$_{\pm 0.9}$ & 99$_{\pm 2.0}$ & \textbf{100$_{\pm 0.0}$} & 99$_{\pm 2.0}$ & 94$_{\pm 3.7}$ & 96$_{\pm 3.7}$ & 95$_{\pm 3.2}$ & 79$_{\pm 4.9}$ & 89$_{\pm 9.7}$ & 84$_{\pm 10.2}$ \\ 
BridgeVLA~\citep{bridgevla} & 65.0$_{\pm 1.3}$ & \textbf{100}$_{\pm 0.0}$ & \textbf{100}$_{\pm 0.0}$ & \textbf{100}$_{\pm 0.0}$ & 74$_{\pm 9.7}$ & 89$_{\pm 4.9}$ & 0$_{\pm 0.0}$ & 91$_{\pm 3.7}$ & 90$_{\pm 3.2}$ & 90$_{\pm 6.3}$ \\
\rowcolor[HTML]{EFEFEF}
\method & \textbf{83.2}$_{\pm 1.9}$ & \textbf{100}$_{\pm 0.0}$ & \textbf{100}$_{\pm 0.0}$ & \textbf{100}$_{\pm 0.0}$ & \tb{99}$_{\pm 2.2}$ & \tb{100}$_{\pm 0.0}$ & \tb{98}$_{\pm 2.7}$ & \textbf{93}$_{\pm 4.5}$ & \textbf{97}$_{\pm 2.7}$ & \textbf{98}$_{\pm 2.7}$ \\

\midrule

\rowcolor[HTML]{CBCEFB}
Method & \begin{tabular}[c]{@{}c@{}}Stack\\ Blocks+24\end{tabular} & \begin{tabular}[c]{@{}c@{}}Stack\\ Blocks+27\end{tabular} & \begin{tabular}[c]{@{}c@{}}Stack\\ Blocks+33\end{tabular} & \begin{tabular}[c]{@{}c@{}}Slide\\ Block+2\end{tabular} & \begin{tabular}[c]{@{}c@{}}Slide\\ Block+3\end{tabular} & \begin{tabular}[c]{@{}c@{}}Close\\ Jar+3\end{tabular} & \begin{tabular}[c]{@{}c@{}}Close\\ Jar+4\end{tabular} & \begin{tabular}[c]{@{}c@{}}LightBulb\\ In+1\end{tabular} & \begin{tabular}[c]{@{}c@{}}LightBulb\\ In+2\end{tabular} & \begin{tabular}[c]{@{}c@{}}Lamp\\ On+0\end{tabular} \\ \midrule
HiveFormer~\citep{hiveformer} & 0$_{\pm 0.0}$ & 4$_{\pm 4.2}$ & 0$_{\pm 0.0}$ & 0$_{\pm 0.0}$ & 0$_{\pm 0.0}$ & 0$_{\pm 0.0}$ & 0$_{\pm 0.0}$ & 4$_{\pm 4.2}$ & 0$_{\pm 0.0}$ & 7$_{\pm 4.5}$ \\
\rowcolor[HTML]{EFEFEF}
PolarNet~\citep{polarnet} & 1$_{\pm 2.2}$ & 2$_{\pm 2.7}$ & 6$_{\pm 8.2}$ & 0$_{\pm 0.0}$ & 0$_{\pm 0.0}$ & 20$_{\pm 10.6}$ & 82$_{\pm 5.7}$ & 22$_{\pm 11.5}$ & 17$_{\pm 8.4}$ & 14$_{\pm 10.8}$ \\
3D Diffuser Actor~\citep{3ddiffuseractor} & 66$_{\pm 13.9}$ & 82$_{\pm 2.7}$ & 50$_{\pm 14.6}$ & 0$_{\pm 0.0}$ & 0$_{\pm 0.0}$ & 23$_{\pm 16.8}$ & 82$_{\pm 5.7}$ & 51$_{\pm 17.8}$ & 60$_{\pm 10.0}$ & 7$_{\pm 7.6}$ \\
\rowcolor[HTML]{EFEFEF}
RVT2~\citep{rvt2} & 18$_{\pm 4.5}$ & 56$_{\pm 16.7}$ & 45$_{\pm 13.7}$ & 0$_{\pm 0.0}$ & 1$_{\pm 2.2}$ & 7$_{\pm 7.6}$ & 77$_{\pm 5.7}$ & \textbf{68}$_{\pm 14.4}$ & 6$_{\pm 6.5}$ & 0$_{\pm 0.0}$ \\
3D-LOTUS~\citep{gembench} & 13$_{\pm 8.1}$ & 40$_{\pm 9.5}$ & 69$_{\pm 5.8}$ & 0$_{\pm 0.0}$ & 0$_{\pm 0.0}$ & 71$_{\pm 5.8}$ & 90$_{\pm 4.5}$ & 24$_{\pm 4.9}$ & 41$_{\pm 8.6}$ & 0$_{\pm 0.0}$ \\ 
\rowcolor[HTML]{EFEFEF}
3D-LOTUS++~\citep{gembench} & 22$_{\pm 9.3}$ & 83$_{\pm 7.5}$ & 59$_{\pm 3.7}$ & \textbf{27}$_{\pm 9.8}$ & 5$_{\pm 3.2}$ & \textbf{98}$_{\pm 2.5}$ & 96$_{\pm 3.7}$ & 56$_{\pm 9.7}$ & 43$_{\pm 7.5}$ & 2$_{\pm 2.0}$ \\ 
BridgeVLA~\citep{bridgevla} & 61$_{\pm 10.7}$ & 51$_{\pm 13.2}$ & 79$_{\pm 8.6}$ & 12$_{\pm 9.3}$ & 3$_{\pm 4.0}$ & 66$_{\pm 6.6}$ & 88$_{\pm 4.0}$ & 66$_{\pm 8.6}$ & 74$_{\pm 5.8}$ & 7$_{\pm 4.0}$ \\
\rowcolor[HTML]{EFEFEF}
\method & \tb{95}$_{\pm 3.5}$ & \tb{86}$_{\pm 2.2}$ & \textbf{91}$_{\pm 4.2}$ & 18$_{\pm 5.7}$ & \tb{68}$_{\pm 5.7}$ & 95$_{\pm 3.5}$ & \tb{98}$_{\pm 4.5}$ & 66$_{\pm 5.5}$ & \textbf{81}$_{\pm 6.5}$ & \tb{20}$_{\pm 6.1}$ \\
\midrule

\rowcolor[HTML]{CBCEFB}
Method & \begin{tabular}[c]{@{}c@{}}Reach\&\\ Drag+5\end{tabular} & \begin{tabular}[c]{@{}c@{}}Reach\&\\ Drag+7\end{tabular} & \begin{tabular}[c]{@{}c@{}}PutCube\\ InSafe+0\end{tabular} & \begin{tabular}[c]{@{}c@{}}Pick\&Lift\\ Cylinder+0\end{tabular} & \begin{tabular}[c]{@{}c@{}}Pick\&Lift\\ Star+0\end{tabular} & \begin{tabular}[c]{@{}c@{}}Pick\&Lift\\ Moon+0\end{tabular} & \begin{tabular}[c]{@{}c@{}}Pick\&Lift\\ Toy+0\end{tabular} & \begin{tabular}[c]{@{}c@{}}PutIn\\ Cupboard+7\end{tabular} & \begin{tabular}[c]{@{}c@{}}PutIn\\ Cupboard+8\end{tabular} &  \\ \midrule
HiveFormer~\citep{hiveformer} & 1$_{\pm 2.2}$ & 0$_{\pm 0.0}$ & 4$_{\pm 2.2}$ & 78$_{\pm 5.7}$ & 73$_{\pm 7.6}$ & 88$_{\pm 2.7}$ & 87$_{\pm 4.5}$ & 0$_{\pm 0.0}$ & 0$_{\pm 0.0}$ &  \\
\rowcolor[HTML]{EFEFEF}
PolarNet~\citep{polarnet} & 61$_{\pm 8.2}$ & 10$_{\pm 6.1}$ & 40$_{\pm 14.1}$ & 93$_{\pm 6.7}$ & 88$_{\pm 8.4}$ & 93$_{\pm 6.7}$ & 90$_{\pm 3.5}$ & 0$_{\pm 0.0}$ & 0$_{\pm 0.0}$ &  \\
3D Diffuser Actor~\citep{3ddiffuseractor} & 0$_{\pm 0.0}$ & 64$_{\pm 6.5}$ & 3$_{\pm 2.7}$ & \textbf{99}$_{\pm 2.2}$ & 43$_{\pm 17.9}$ & 91$_{\pm 9.6}$ & 30$_{\pm 9.4}$ & 0$_{\pm 0.0}$ & 3$_{\pm 4.5}$ &  \\
\rowcolor[HTML]{EFEFEF}
RVT2~\citep{rvt2} & 91$_{\pm 2.2}$ & 89$_{\pm 6.5}$ & 6$_{\pm 5.5}$ & 98$_{\pm 2.7}$ & 98$_{\pm 4.5}$ & 94$_{\pm 4.2}$ & 78$_{\pm 8.4}$ & 0$_{\pm 0.0}$ & 0$_{\pm 0.0}$ &  \\
3D-LOTUS~\citep{gembench} & \textbf{95}$_{\pm 4.5}$ & 18$_{\pm 10.8}$ & 25$_{\pm 5.5}$ & 88$_{\pm 8.7}$ & 69$_{\pm 6.6}$ & 80$_{\pm 8.4}$ & \textbf{96}$_{\pm 3.7}$ & 0$_{\pm 0.0}$ & 0$_{\pm 0.0}$  & \\ 
\rowcolor[HTML]{EFEFEF}
3D-LOTUS++~\citep{gembench}  & 94$_{\pm 2.0}$ & 64$_{\pm 12.4}$ & 37$_{\pm 5.1}$ & 91$_{\pm 2.0}$ & 94$_{\pm 3.7}$ & 29$_{\pm 6.6}$ & 71$_{\pm 2.0}$ & 1$_{\pm 2.0}$ & 0$_{\pm 0.0}$  & \\ 
BridgeVLA~\citep{bridgevla} & 94$_{\pm 3.7}$ & \textbf{96}$_{\pm 3.7}$ & 3$_{\pm 2.5}$ & 98$_{\pm 2.5}$ & 99$_{\pm 2.0}$ & 95$_{\pm 3.2}$ & 93$_{\pm 5.1}$ & 0$_{\pm 0.0}$ & 0$_{\pm 0.0}$ & \\
\rowcolor[HTML]{EFEFEF}
\method & \tb{95}$_{\pm 3.5}$ & 90$_{\pm 6.1}$ & \tb{61}$_{\pm 14.7}$ & 97$_{\pm 2.7}$ & \textbf{100}$_{\pm 0.0}$ & \textbf{98}$_{\pm 2.7}$ & 84$_{\pm 8.2}$ & \tb{50}$_{\pm 11.2}$ & \tb{53}$_{\pm 12.5}$ & \\
\bottomrule
\end{tabular}
}
\caption{\textbf{Per-task Success Rate on GemBench Level 2.}}
\label{tab:gembench_l2_all}
\end{table*}

\begin{table*}
\centering
\resizebox{\textwidth}{!}{ 
\begin{tabular}{lcccccccc} \toprule
\rowcolor[HTML]{CBCEFB}
Method & Avg. & \begin{tabular}[c]{@{}c@{}}Close\\ Door+0\end{tabular} & \begin{tabular}[c]{@{}c@{}}Close\\ Box+0\end{tabular} & \begin{tabular}[c]{@{}c@{}}Close\\ Fridge2+0\end{tabular} & \begin{tabular}[c]{@{}c@{}}CloseLaptop\\ Lid2+0\end{tabular} & \begin{tabular}[c]{@{}c@{}}Close\\ Microwave2+0\end{tabular} & \begin{tabular}[c]{@{}c@{}}Open\\ Door2+0\end{tabular} & \begin{tabular}[c]{@{}c@{}}Open\\ Box2+0\end{tabular} \\
HiveFormer~\citep{hiveformer} & 35.1$_{\pm 1.7}$ & 0$_{\pm 0.0}$ & 1$_{\pm 2.2}$ & 34$_{\pm 9.6}$ & 52$_{\pm 9.1}$ & 15$_{\pm 7.1}$ & 32$_{\pm 11.5}$ & 5$_{\pm 3.5}$ \\
\rowcolor[HTML]{EFEFEF}
PolarNet~\citep{polarnet} & 38.5$_{\pm 1.7}$ & 0$_{\pm 0.0}$ & 0$_{\pm 0.0}$ & 78$_{\pm 5.7}$ & 26$_{\pm 8.2}$ & 74$_{\pm 6.5}$ & 33$_{\pm 6.7}$ & 23$_{\pm 8.4}$ \\
3D Diffuser Actor~\citep{3ddiffuseractor} & 37.0$_{\pm 2.2}$ & 0$_{\pm 0.0}$ & 0$_{\pm 0.0}$ & \textbf{97}$_{\pm 2.7}$ & 23$_{\pm 6.7}$ & 88$_{\pm 7.6}$ & \textbf{86}$_{\pm 7.4}$ & 67$_{\pm 9.8}$ \\
\rowcolor[HTML]{EFEFEF}
RVT2~\citep{rvt2} & 36.0$_{\pm 2.2}$ & 1$_{\pm 2.2}$ & 2$_{\pm 2.7}$ & 72$_{\pm 6.7}$ & 42$_{\pm 14.0}$ & 71$_{\pm 8.9}$ & 79$_{\pm 6.5}$ & 5$_{\pm 6.1}$ \\
3D-LOTUS~\citep{gembench} & 38.1$_{\pm 1.1}$ & 0$_{\pm 0.0}$ & \textbf{58}$_{\pm 8.1}$ & 36$_{\pm 9.7}$ & 54$_{\pm 10.7}$ & 85$_{\pm 7.1}$ & 42$_{\pm 6.8}$ & 11$_{\pm 6.6}$ \\
\rowcolor[HTML]{EFEFEF}
3D-LOTUS++~\citep{gembench} & 41.5$_{\pm 1.8}$ & 1$_{\pm 2.0}$ & 29$_{\pm 8.6}$ & 93$_{\pm 2.5}$ & 50$_{\pm 9.5}$ & \textbf{99}$_{\pm 2.0}$ & 52$_{\pm 10.3}$ & 16$_{\pm 8.0}$ \\
BridgeVLA~\citep{bridgevla} & \textbf{43.8}$_{\pm 1.2}$ & 0$_{\pm 0.0}$ & 1$_{\pm 2.0}$ & 95$_{\pm 5.5}$ & \textbf{77}$_{\pm 4.0}$ & 54$_{\pm 10.2}$ & 68$_{\pm 10.8}$ & \textbf{74}$_{\pm 4.9}$ \\
\rowcolor[HTML]{EFEFEF}
\method & \textbf{49.6}$_{\pm 2.1}$ & \tb{3}$_{\pm 2.7}$ & 9$_{\pm 5.5}$ & 92$_{\pm 4.5}$ & 35$_{\pm 9.4}$ & 79$_{\pm 5.5}$ & 56$_{\pm 6.5}$ & 1$_{\pm 2.2}$ \\

\midrule
\rowcolor[HTML]{CBCEFB}
Method & \begin{tabular}[c]{@{}c@{}}Open\\ Drawer2+0\end{tabular} & \begin{tabular}[c]{@{}c@{}}Open\\ Drawer3+0\end{tabular} & \begin{tabular}[c]{@{}c@{}}OpenDrawer\\ Long+0\end{tabular} & \begin{tabular}[c]{@{}c@{}}OpenDrawer\\ Long+1\end{tabular} & \begin{tabular}[c]{@{}c@{}}OpenDrawer\\ Long+2\end{tabular} & \begin{tabular}[c]{@{}c@{}}OpenDrawer\\ Long+3\end{tabular} & \begin{tabular}[c]{@{}c@{}}Toilet\\ SeatUp+0\end{tabular} & \begin{tabular}[c]{@{}c@{}}Open\\ Fridge+0\end{tabular} \\
HiveFormer~\citep{hiveformer} & 59$_{\pm 11.9}$ & 39$_{\pm 11.9}$ & 78$_{\pm 8.4}$ & 82$_{\pm 4.5}$ & 49$_{\pm 4.2}$ & 57$_{\pm 11.5}$ & 6$_{\pm 4.2}$ & 0$_{\pm 0.0}$ \\
\rowcolor[HTML]{EFEFEF}
PolarNet~\citep{polarnet} & \textbf{91}$_{\pm 4.2}$ & 29$_{\pm 8.2}$ & 84$_{\pm 11.9}$ & 88$_{\pm 5.7}$ & \textbf{63}$_{\pm 8.4}$ & 37$_{\pm 7.6}$ & 2$_{\pm 2.7}$ & 4$_{\pm 2.2}$ \\
3D Diffuser Actor~\citep{3ddiffuseractor} & 19$_{\pm 8.2}$ & 1$_{\pm 2.2}$ & 15$_{\pm 5.0}$ & 35$_{\pm 13.7}$ & 26$_{\pm 9.6}$ & 79$_{\pm 12.9}$ & 0$_{\pm 0.0}$ & \textbf{7}$_{\pm 5.7}$ \\
\rowcolor[HTML]{EFEFEF}
RVT2~\citep{rvt2} & 81$_{\pm 11.9}$ & 0$_{\pm 0.0}$ & \textbf{84}$_{\pm 8.2}$ & 39$_{\pm 10.8}$ & 11$_{\pm 8.9}$ & 75$_{\pm 6.1}$ & 7$_{\pm 5.7}$ & 0$_{\pm 0.0}$ \\
3D-LOTUS~\citep{gembench} & 90$_{\pm 3.2}$ & 22$_{\pm 8.1}$ & 56$_{\pm 13.9}$ & 33$_{\pm 11.2}$ & 17$_{\pm 8.1}$ & 75$_{\pm 6.3}$ & 0$_{\pm 0.0}$ & 4$_{\pm 5.8}$ \\
\rowcolor[HTML]{EFEFEF}
3D-LOTUS++~\citep{gembench} & 70$_{\pm 5.5}$ & 41$_{\pm 4.9}$ & 72$_{\pm 4.0}$ & 52$_{\pm 10.8}$ & 23$_{\pm 8.1}$ & 78$_{\pm 5.1}$ & \textbf{8}$_{\pm 5.1}$ & 0$_{\pm 0.0}$ \\
BridgeVLA~\citep{bridgevla} & 65$_{\pm 6.3}$ & \textbf{87}$_{\pm 6.0}$ & 59$_{\pm 8.6}$ & 34$_{\pm 8.0}$ & 18$_{\pm 10.3}$ & \textbf{85}$_{\pm 8.4}$ & 6$_{\pm 5.8}$ & \textbf{7}$_{\pm 2.5}$ \\
\method & 68$_{\pm 8.4}$ & \textbf{87}$_{\pm 7.6}$ & 44$_{\pm 10.8}$ & \tb{94}$_{\pm 5.5}$ & 14$_{\pm 5.5}$ & 76$_{\pm 13.4}$ & 7$_{\pm 4.5}$ & 3$_{\pm 4.5}$ \\
\midrule

\rowcolor[HTML]{CBCEFB}
Method & \begin{tabular}[c]{@{}c@{}}OpenLaptop\\ Lid+0\end{tabular} & \begin{tabular}[c]{@{}c@{}}Open\\ Microwave+0\end{tabular} & \begin{tabular}[c]{@{}c@{}}PutMoney\\ InSafe+2\end{tabular} & \begin{tabular}[c]{@{}c@{}}Open\\ Drawer+1\end{tabular} & \begin{tabular}[c]{@{}c@{}}Close\\ Drawer+0\end{tabular} & \begin{tabular}[c]{@{}c@{}}Close\\ Grill+0\end{tabular} &  &  \\
HiveFormer~\citep{hiveformer} & \textbf{100}$_{\pm 0.0}$ & 0$_{\pm 0.0}$ & 0$_{\pm 0.0}$ & 0$_{\pm 0.0}$ & 83$_{\pm 5.7}$ & 44$_{\pm 10.8}$ &  &  \\
\rowcolor[HTML]{EFEFEF}
PolarNet~\citep{polarnet} & \textbf{100}$_{\pm 0.0}$ & 0$_{\pm 0.0}$ & 1$_{\pm 2.2}$ & 4$_{\pm 4.2}$ & 29$_{\pm 11.9}$ & 42$_{\pm 11.5}$ &  &  \\
3D Diffuser Actor~\citep{3ddiffuseractor} & \textbf{100}$_{\pm 0.0}$ & 0$_{\pm 0.0}$ & 2$_{\pm 4.5}$ & 0$_{\pm 0.0}$ & 66$_{\pm 7.4}$ & \textbf{65}$_{\pm 13.7}$ &  &  \\
\rowcolor[HTML]{EFEFEF}
RVT2~\citep{rvt2} & 93$_{\pm 5.7}$ & 0$_{\pm 0.0}$ & 0$_{\pm 0.0}$ & 6$_{\pm 2.2}$ & 78$_{\pm 8.4}$ & 9$_{\pm 4.2}$ &  &  \\
3D-LOTUS~\citep{gembench} & \textbf{100}$_{\pm 0.0}$ & 0$_{\pm 0.0}$ & 0$_{\pm 0.0}$ & 0$_{\pm 0.0}$ & \textbf{87}$_{\pm 8.1}$ & 29$_{\pm 6.6}$ &  &  \\
\rowcolor[HTML]{EFEFEF}
3D-LOTUS++~\citep{gembench} & 86$_{\pm 6.6}$ & 0$_{\pm 0.0}$ & 13$_{\pm 8.1}$ & 0$_{\pm 0.0}$ & 69$_{\pm 5.8}$ & 19$_{\pm 13.9}$ &  &  \\
BridgeVLA~\citep{bridgevla} & 95$_{\pm 0.0}$ & 0$_{\pm 0.0}$ & 2$_{\pm 2.5}$ & 0$_{\pm 0.0}$ & 58$_{\pm 12.9}$ & 35$_{\pm 12.3}$ &  &  \\
\rowcolor[HTML]{EFEFEF}
\method & 78$_{\pm 9.1}$ & 0$_{\pm 0.0}$ & \tb{76}$_{\pm 5.5}$ & \tb{96}$_{\pm 6.5}$ & 84$_{\pm 8.2}$ & 40$_{\pm 7.1}$ &  &  \\
\bottomrule
\end{tabular}
}
\caption{\tb{Per-task Success Rate on GemBench Level 3.}}
\label{tab:gembench_l3_all}
\end{table*}

\begin{table*}[t]
\centering
\resizebox{\textwidth}{!}{ 
\begin{tabular}{lccccccc} \toprule
\rowcolor[HTML]{CBCEFB}
Method & Avg. & \begin{tabular}[c]{@{}c@{}}Push\\ Buttons4+1\end{tabular} & \begin{tabular}[c]{@{}c@{}}Push\\ Buttons4+2\end{tabular} & \begin{tabular}[c]{@{}c@{}}Push\\ Buttons4+3\end{tabular} & \begin{tabular}[c]{@{}c@{}}TakeShoes\\ OutOfBox+0\end{tabular} & \begin{tabular}[c]{@{}c@{}}PutItems\\ InDrawer+0\end{tabular} & \begin{tabular}[c]{@{}c@{}}PutItems\\ InDrawer+2\end{tabular} \\
HiveFormer~\citep{hiveformer} & 0$_{\pm 0.0}$ & 0$_{\pm 0.0}$ & 0$_{\pm 0.0}$ & 0$_{\pm 0.0}$ & 0$_{\pm 0.0}$ & 0$_{\pm 0.0}$ & 0$_{\pm 0.0}$ \\
\rowcolor[HTML]{EFEFEF}
PolarNet~\citep{polarnet} & 0.1$_{\pm 0.2}$ & 1$_{\pm 2.2}$ & 0$_{\pm 0.0}$ & 0$_{\pm 0.0}$ & 0$_{\pm 0.0}$ & 0$_{\pm 0.0}$ & 0$_{\pm 0.0}$ \\
3D Diffuser Actor~\citep{3ddiffuseractor} & 0$_{\pm 0.0}$ & 0$_{\pm 0.0}$ & 0$_{\pm 0.0}$ & 0$_{\pm 0.0}$ & 0$_{\pm 0.0}$ & 0$_{\pm 0.0}$ & 0$_{\pm 0.0}$ \\
\rowcolor[HTML]{EFEFEF}
RVT2~\citep{rvt2} & 0$_{\pm 0.0}$ & 0$_{\pm 0.0}$ & 0$_{\pm 0.0}$ & 0$_{\pm 0.0}$ & 0$_{\pm 0.0}$ & 0$_{\pm 0.0}$ & 0$_{\pm 0.0}$ \\
3D-LOTUS~\citep{gembench} & 0.3$_{\pm 0.3}$ & 3$_{\pm 4.0}$ & 0$_{\pm 0.0}$ & 0$_{\pm 0.0}$ & 0$_{\pm 0.0}$ & 0$_{\pm 0.0}$ & 0$_{\pm 0.0}$ \\
\rowcolor[HTML]{EFEFEF}
3D-LOTUS++~\citep{gembench} & 17.4$_{\pm 0.4}$ & 76$_{\pm 7.4}$ & 49$_{\pm 8.6}$ & 37$_{\pm 8.1}$ & 0$_{\pm 0.0}$ & 0$_{\pm 0.0}$ & 0$_{\pm 0.0}$ \\
BridgeVLA~\citep{bridgevla} & 0$_{\pm 0.0}$ & 0$_{\pm 0.0}$ & 0$_{\pm 0.0}$ & 0$_{\pm 0.0}$ & 0$_{\pm 0.0}$ & 0$_{\pm 0.0}$ & 0$_{\pm 0.0}$ \\
\rowcolor[HTML]{EFEFEF}
\method & \tb{31.4}$_{\pm 0.6}$ & \tb{98}$_{\pm 2.7}$ & \tb{87}$_{\pm 4.5}$ & \tb{92}$_{\pm 5.7}$ & 0$_{\pm 0.0}$ & 0$_{\pm 0.0}$ & 0$_{\pm 0.0}$ \\

\midrule

\rowcolor[HTML]{CBCEFB}
Method & \begin{tabular}[c]{@{}c@{}}PutItems\\ InDrawer+4\end{tabular} & Tower4+1 & Tower4+3 & \begin{tabular}[c]{@{}c@{}}Stack\\ Cups+0\end{tabular} & \begin{tabular}[c]{@{}c@{}}Stack\\ Cups+3\end{tabular} & \begin{tabular}[c]{@{}c@{}}PutAllGroceries\\ InCupboard+0\end{tabular} &  \\
HiveFormer~\citep{hiveformer} & 0$_{\pm 0.0}$ & 0$_{\pm 0.0}$ & 0$_{\pm 0.0}$ & 0$_{\pm 0.0}$ & 0$_{\pm 0.0}$ & 0$_{\pm 0.0}$ &  \\
\rowcolor[HTML]{EFEFEF}
PolarNet~\citep{polarnet} & 0$_{\pm 0.0}$ & 0$_{\pm 0.0}$ & 0$_{\pm 0.0}$ & 0$_{\pm 0.0}$ & 0$_{\pm 0.0}$ & 0$_{\pm 0.0}$ &  \\
3D Diffuser Actor~\citep{3ddiffuseractor} & 0$_{\pm 0.0}$ & 0$_{\pm 0.0}$ & 0$_{\pm 0.0}$ & 0$_{\pm 0.0}$ & 0$_{\pm 0.0}$ & 0$_{\pm 0.0}$ &  \\
\rowcolor[HTML]{EFEFEF}
RVT2~\citep{rvt2} & 0$_{\pm 0.0}$ & 0$_{\pm 0.0}$ & 0$_{\pm 0.0}$ & 0$_{\pm 0.0}$ & 0$_{\pm 0.0}$ & 0$_{\pm 0.0}$ &  \\
3D-LOTUS~\citep{gembench} & 0$_{\pm 0.0}$ & 0$_{\pm 0.0}$ & 0$_{\pm 0.0}$ & 0$_{\pm 0.0}$ & 0$_{\pm 0.0}$ & 0$_{\pm 0.0}$ &  \\
\rowcolor[HTML]{EFEFEF}
3D-LOTUS++~\citep{gembench} & 0$_{\pm 0.0}$ & 17$_{\pm 10.8}$ & 30$_{\pm 13.4}$ & 0$_{\pm 0.0}$ & 0$_{\pm 0.0}$ & 0$_{\pm 0.0}$ & \\
BridgeVLA~\citep{bridgevla} & 0$_{\pm 0.0}$ & 0$_{\pm 0.0}$ & 0$_{\pm 0.0}$ & 0$_{\pm 0.0}$ & 0$_{\pm 0.0}$ & 0$_{\pm 0.0}$ & \\
\rowcolor[HTML]{EFEFEF}
\method & 0$_{\pm 0.0}$ & \tb{30}$_{\pm 7.1}$ & \tb{70}$_{\pm 6.1}$ & 0$_{\pm 0.0}$ & 0$_{\pm 0.0}$ & 0$_{\pm 0.0}$ & \\
\bottomrule
\end{tabular}
}
\caption{\tb{Per-task Success Rate on GemBench Level 4.}}
\label{tab:gembench_l4_all}
\end{table*}

The results for each task across different generalization levels in Gembench are listed in \Cref{tab:gembench_l1_all}, \Cref{tab:gembench_l2_all},
\Cref{tab:gembench_l3_all},
\Cref{tab:gembench_l4_all}.

\revise{
\subsection{Additional Ablation Study}
\label{appendix:additional ablation}
We first show the performance of our method on GemBench trained with different numbers of robot trajectories (10, 20, 50, 100) in \Cref{tab:number of traj}.
Further increasing the number of robot trajectory improves on the in-domain performance (L1) while does not help in the average success rate.
We conclude that the data efficiency is attributed to the following points.
    \begin{itemize}
        \item Task decomposition allows the agent to learn common skills among different tasks and allows compositional generalization.
        \item Bridging VLM and 3D manipulation with projected images, separating task planning from keypoint prediction, and Chain-of-Thought reasoning lead to an effective adaptation of pre-trained VLM to the coarse task planner.
        \item 
        Data efficiency with respect to the robot trajectories also benefits from leveraging language plans and object position data.
        By co-training on these additional sources, our framework equips the model with task-planning and object-grounding abilities, leaving it to learn only how to act given sub-task instructions and known object locations.
        In contrast, prior approaches required the model to learn task planning, object detection, and action prediction solely from robot trajectories.
    \end{itemize}

\begin{table}[t]
\centering
\caption{\textbf{Ablation on number of robot demonstrations used for training.}
We compare the results of training our method with different numbers of demonstrations.
The results are recorded with three runs of different random seeds.}
\label{tab:number of traj}
\resizebox{\textwidth}{!}{
\begin{tabular}{cccccc}
                                 Number of Demos        & L1 & L2 & L3 & L4 & Average \\ \hline
10                                      &  $84.5 \pm 0.8$ & $81.5 \pm 0.6$   & $43.3 \pm 1.9$   & $30.5 \pm 2.1$   &  $60.0 \pm 0.1$       \\
20  & $83.9 \pm 0.3$   & $83.2 \pm 1.9$   & $49.6 \pm 2.1$   & $31.4 \pm 0.6$   &    \textbf{$62.0 \pm 0.5$}   \\ 
50                                & $87.9 \pm 0.3$   & $81.2 \pm 0.1$  & $47.0 \pm 1.5$  &  $28.3 \pm 2.2$  & $61.1 \pm 0.9$        \\
100 & $86.9 \pm 1.5$   & $81.7 \pm 0.4$   & $47.8 \pm 2.6$   & $27.5 \pm 1.2$   &    $61.0 \pm 0.7$  \\   
\end{tabular}
}
\end{table}

We then show the results of variations on inputs to the fine-grained action predictor in \Cref{tab:ablation on inputs to action predictor}.
Removing some components in the fine-grained action predictor leads to performance drop to some extend. 

\begin{table}[h]
\centering
\caption{\textbf{Ablation on inputs to Fine-grained Action Predictor.}
We compare the results of removing some inputs to our fine-grained action predictor.
The results are recorded with three runs of different random seeds.}
\label{tab:ablation on inputs to action predictor}
\resizebox{\textwidth}{!}{
\begin{tabular}{cccccc}
                                Inputs         & L1 & L2 & L3 & L4 & Average \\ \hline
RGB                                      &  $83.9 \pm 1.4$ & $79.7 \pm 2.3$   & $48.2 \pm 1.3$   & $31.8 \pm 1.3$   &  $60.9 \pm 0.6$       \\
RGB+Depth                                & $87.2 \pm 2.3$   & $83.4 \pm 1.0$  & $48.2 \pm 1.0$  &  $24.3 \pm 0.9$  & $60.8 \pm 0.6$        \\
RGB+Depth+3D positional embedding (Ours) & $83.9 \pm 0.3$   & $83.2 \pm 1.9$   & $49.6 \pm 2.1$   & $31.4 \pm 0.6$   &    \textbf{$62.0 \pm 0.5$}    
\end{tabular}
}
\end{table}

}

\subsection{Real-world Experiments}
\label{appendix:real-world exp evaluation}
\begin{figure}
    \centering
    \includegraphics[width=0.99\linewidth]{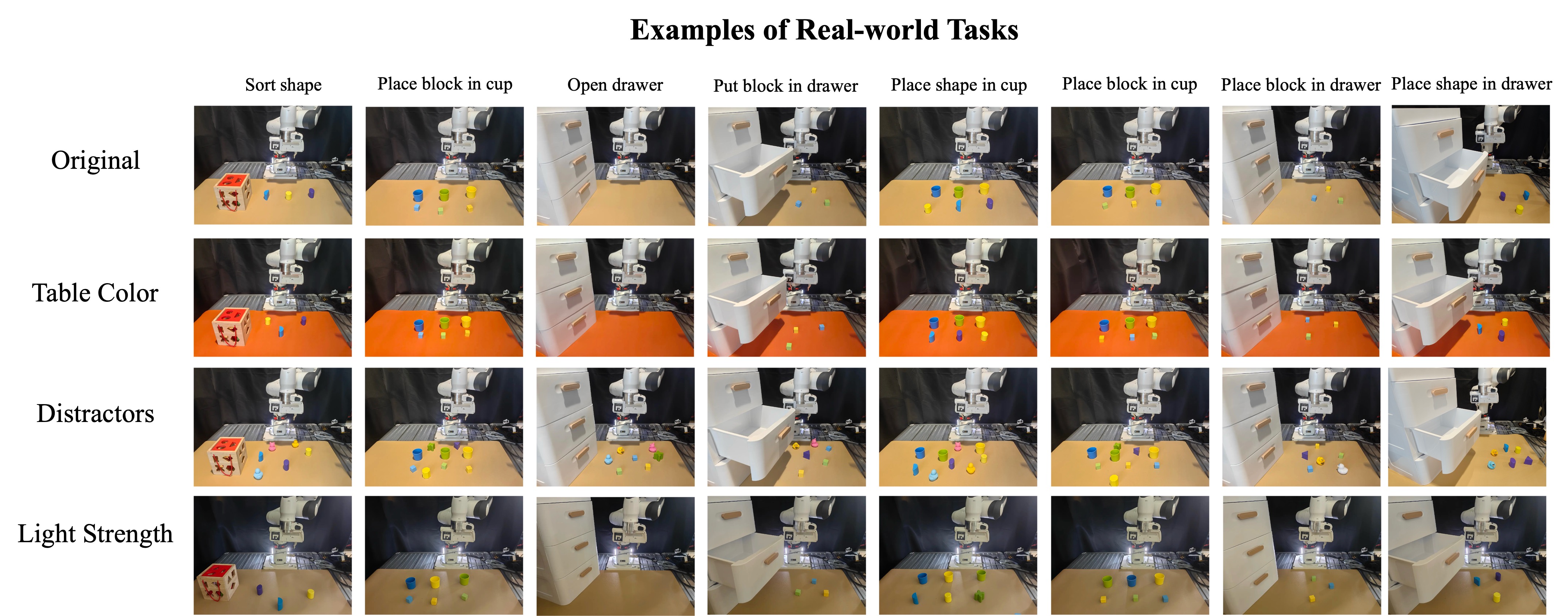}
    \caption{\tb{Overview of the evaluation tasks in real-world experiments.}
    We evaluate the all these eight tasks acroos different variations and record the success rate.}
    \label{fig:all_real_world_evaluation_tasks}
\end{figure}

An overview of the tasks used in real-world experiments are shown in \Cref{fig:all_real_world_evaluation_tasks}.

\end{document}